\documentclass[sigconf,nonacm]{acmart}
\usepackage{threeparttable}
\usepackage{multirow}
\usepackage{tikz}
\usepackage{pgfplots}
\usepackage{pgfplotstable}
\usepackage{xcolor}
\usepackage{subcaption}
\usepackage{enumitem}
\usepackage{float}
\usepackage{stfloats}
\usepackage{diagbox}
\usepackage{url}

\pgfplotsset{compat=1.18}
    \def\addlegendimage{\csname pgfplots@addlegendimage\endcsname}

\definecolor{mae}{HTML}{32012F}
\definecolor{rmse}{HTML}{F97300}

\AtBeginDocument{%
  \providecommand\BibTeX{{%
    \normalfont B\kern-0.5em{\scshape i\kern-0.25em b}\kern-0.8em\TeX}}}

\setcopyright{acmlicensed}
\copyrightyear{2018}
\acmYear{2018}
\acmDOI{XXXXXXX.XXXXXXX}

\acmConference[Conference acronym 'XX]{Make sure to enter the correct
  conference title from your rights confirmation emai}{June 03--05,
  2018}{Woodstock, NY}
\acmISBN{978-1-4503-XXXX-X/18/06}

\newtheorem{definition}{Definition}

\begin{document}

\title{TrajFM: A Vehicle Trajectory Foundation Model for Region and Task Transferability}

\author{Yan Lin}
\authornote{Both authors contributed equally to this research.}
\email{lyan@cs.aau.dk}
\affiliation{%
  \institution{Aalborg University}
  \city{Aalborg}
  \country{Denmark}
}

\author{Tonglong Wei}
\authornotemark[1]
\author{Zeyu Zhou}
\email{{weitonglong, zeyuzhou}@bjtu.edu.cn}
\affiliation{%
  \institution{Beijing Jiaotong University}
  \city{Beijing}
  \country{China}
}

\author{Haomin Wen}
\email{haominwe@andrew.cmu.edu}
\affiliation{%
  \institution{Carnegie Mellon University}
  \city{Pittsburgh}
  \country{USA}
}

\author{Jilin Hu}
\email{jlhu@dase.ecnu.edu.cn}
\affiliation{%
  \institution{East China Normal University}
  \city{Shanghai}
  \country{China}
}

\author{Shengnan Guo}
\author{Youfang Lin}
\email{{guoshn, yflin}@bjtu.edu.cn}
\affiliation{%
  \institution{Beijing Jiaotong University}
  \city{Beijing}
  \country{China}
}

\author{Huaiyu Wan}
\authornote{Corresponding author.}
\email{hywan@bjtu.edu.cn}
\affiliation{%
  \institution{Beijing Jiaotong University}
  \city{Beijing}
  \country{China}
}

\renewcommand{\shortauthors}{Yan Lin, Tonglong Wei, et al.}

\begin{abstract}
Vehicle trajectories provide valuable movement information that supports various downstream tasks and powers real-world applications. A desirable trajectory learning model should transfer between different regions and tasks without retraining, thus improving computational efficiency and effectiveness with limited training data. However, a model's ability to transfer across regions is limited by the unique spatial features and POI arrangements of each region, which are closely linked to vehicle movement patterns and difficult to generalize. Additionally, achieving task transferability is challenging due to the differing generation schemes required for various tasks. Existing efforts towards transferability primarily involve learning embedding vectors for trajectories, which perform poorly in region transfer and still require retraining of prediction modules for task transfer.

To address these challenges, we propose TrajFM, a vehicle trajectory foundation model that excels in both region and task transferability. For region transferability, we introduce STRFormer as the main learnable model within TrajFM. It integrates spatial, temporal, and POI modalities of trajectories to effectively manage variations in POI arrangements across regions and includes a learnable spatio-temporal Rotary position embedding module for handling spatial features. For task transferability, we propose a trajectory masking and recovery scheme. This scheme unifies the generation processes of various tasks into the masking and recovery of modalities and sub-trajectories, allowing TrajFM to be pre-trained once and transferred to different tasks without retraining. Experiments on two real-world vehicle trajectory datasets under various settings demonstrate the effectiveness of TrajFM.
Code is available at \url{https://anonymous.4open.science/r/TrajFM-30E4}.
\end{abstract}

\maketitle

\section{Introduction}
\label{sec:introduction}
A vehicle trajectory is a sequence of (location, time) pairs that record a vehicle's movement during travel. With the widespread adoption of location recording devices, such as in-vehicle navigation systems and smartphones, vehicle trajectories are becoming increasingly available. This availability, combined with the recent emphasis on intelligent traffic systems in many regions, allows vehicle trajectories to offer valuable movement information. They find important applications in various real-world scenarios, such as movement prediction~\cite{DBLP:conf/ijcai/WuCSZW17,DBLP:journals/dase/YuanL21,DBLP:journals/www/YanZSYD23}, anomaly detection~\cite{DBLP:conf/icde/Liu0CB20,DBLP:journals/pvldb/HanCMG22,DBLP:conf/icde/Zhang0LHYLCS23}, trajectory-user linkage~\cite{DBLP:conf/cikm/LiangOWLCZZZ22,DBLP:journals/www/SangXCZ23,DBLP:journals/tkdd/ChenHYJD24}, and travel time estimation~\cite{DBLP:conf/aaai/WuW19,DBLP:conf/sigmod/Yuan0BF20,DBLP:journals/pacmmod/LinWHGYLJ23}. The increased availability and utilization of vehicle trajectories encourage the development of trajectory learning models that can perform various tasks to power real-world applications.

\begin{figure}
    \centering
    \includegraphics[width=1.0\linewidth]{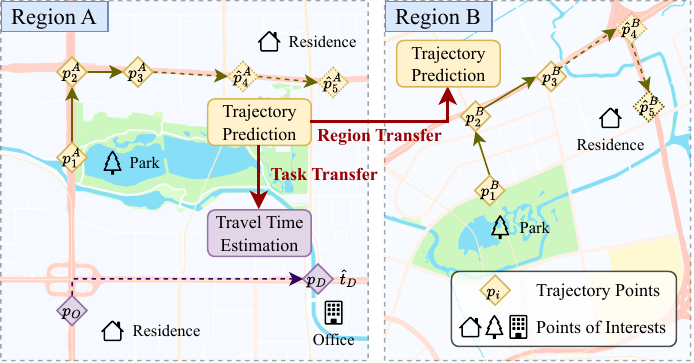}
    \caption{Examples of region and task transferability.}
    \label{fig:motivation}
\end{figure}

Given the widespread use of vehicle trajectories and trajectory learning models across different regions and tasks, it is desirable for a trajectory learning model to have region and task transferability. This means it should be able to transfer between different regions and tasks without retraining. Figure~\ref{fig:motivation} shows an example where a trajectory learning model capable of predicting trajectories in region A is also transferable to the same task in region B and the travel time estimation task. This transferability has two key benefits. First, it allows one universal model to be used in various scenarios without retraining and storing multiple instances, improving overall computational efficiency. Second, transferring a trained model to regions and tasks with limited training data can be more effective than training from scratch in these regions and tasks. Despite these advantages, existing efforts have failed to develop a trajectory learning model with effective region and task transferability due to several challenges.

\textbf{The challenge in region transferability} arises from the differences in spatial features and point of interest (POI) arrangements between regions, leading to significant variations in vehicle movement patterns. For example, consider Figure~\ref{fig:motivation}, which shows two vehicle trajectories: $\mathcal{T}_A = \langle p_1^A, p_2^A, \dots, p_5^A \rangle$ in region A and $\mathcal{T}_B = \langle p_1^B, p_2^B, \dots, p_5^B \rangle$ in region B. The spatial features of these trajectories differ because the regions encompass distinct spatial coordinate ranges. Therefore, a model trained on the spatial features of $\mathcal{T}_A$ cannot be directly applied to $\mathcal{T}_B$. Additionally, both trajectories travel from parks to residences, indicating a shared purpose of returning home after visiting parks. However, due to the different POI arrangements in the two regions, $\mathcal{T}_A$ and $\mathcal{T}_B$ exhibit different movement patterns, preventing the direct transfer of movement information learned from region A to region B.

Existing methods typically employ normalization~\cite{DBLP:conf/ijcnn/YaoZZHB17,DBLP:journals/tkde/WanLGL22}, grid discretization~\cite{DBLP:conf/icde/LiZCJW18,DBLP:journals/www/YanZSYD23}, or map-matching~\cite{DBLP:journals/tist/FuL20,DBLP:conf/icde/JiangPRJLW23,10517676} on spatial features to accommodate the different coordinate ranges of regions. However, these strategies are often ineffective for transferring information between regions. Because the sizes of regions usually differ, normalization and grid discretization result in different scales of spatial features across regions, which hinders a model's ability to transfer between regions. Moreover, since different regions have different road network structures, map-matching does not contribute to a model's region transferability. Additionally, these strategies fail to account for differences in POI arrangements.

\textbf{The challenge in task transferability} stems from the differences in generation schemes and the correlations learned for various tasks, making it difficult for a model to handle a wide range of tasks simultaneously. Consider the two tasks illustrated in Figure~\ref{fig:motivation}. The trajectory prediction task requires generating two future trajectory points $\langle \hat p_4^A, \hat p_5^A \rangle$ given three historical trajectory points $\langle p_1^A, p_2^A, p_3^A \rangle$, focusing on learning the sequential correlation of trajectory points. The travel time estimation task involves generating the arrival time $\hat t_D$ of a future trajectory from origin $p_O$ to destination $p_D$, emphasizing learning the temporal correlation between origin-destination pairs and travel times. This discrepancy makes it challenging for a model designed for trajectory prediction to effectively transfer to travel time estimation, or vise versa.

Existing efforts towards task transferability mostly adhere to the embedding strategy, which learns trajectory encoders for mapping vehicle trajectories into embedding vectors~\cite{DBLP:journals/tist/FuL20,DBLP:conf/ijcai/LiangOYWTZ21,DBLP:conf/icde/JiangPRJLW23,10375102}. Although these embedding vectors contain transferable movement information of trajectories, they still necessitate prediction modules for adaptation to downstream tasks, which require additional training and storage of parameters. Furthermore, the learned trajectory encoders require the integrity of trajectory features. In other words, they transfer poorly to tasks where large portions of trajectory features are not provided as input, such as the travel time estimation task illustrated in Figure~\ref{fig:motivation}.

To address the above challenges, we propose a novel method called the \textit{\underline{Traj}ectory \underline{F}oundation \underline{M}odel} (\textbf{TrajFM}), which excels in region and task transferability. TrajFM encompasses two core components: STRFormer and a trajectory masking and recovery scheme. STRFormer is the learnable model within TrajFM and enables region transferability. It incorporates spatial, temporal, and POI modalities of trajectories to discern differences in POI arrangements across various regions. Additionally, it includes a learnable spatio-temporal Rotary position embedding module for transferable modeling of spatial features across different regions. The trajectory masking and recovery scheme equips TrajFM with task transferability. It unifies the generation processes of different tasks into the masking and recovery of modalities and sub-trajectories, and also acts as a pre-training scheme to enable TrajFM to transfer to various tasks without re-training.

In summary, the contributions of the paper are as follows:
\begin{itemize}[leftmargin=*]
    \item We propose TrajFM, a new vehicle trajectory foundation model that can transfer across different regions and types of downstream tasks without re-training and storing multiple models.
    \item We introduce STRFormer as the learnable model of TrajFM to enable region transferability. It handles regional differences in spatial features and POI arrangements by incorporating the POI modality and a learnable spatio-temporal Rotary position embedding module.
    \item We introduce a trajectory masking and recovery scheme to facilitate task transferability, unifying the generation processes of different tasks into the masking and recovery of modalities and sub-trajectories.
    \item We conduct extensive experiments on two real-world vehicle trajectory datasets with various settings, showcasing the effectiveness of TrajFM in transferable vehicle trajectory learning.
\end{itemize}

\section{Related Works} 
\label{sec:related-works}
In this section, we first review existing works on vehicle trajectory learning as a direct comparison to our proposed method, and then discuss foundation models in other domains.

\subsection{Vehicle Trajectory Learning Methods}
Vehicle trajectory learning methods extract information from vehicle trajectories to perform tasks for trajectory-related applications. These methods can be broadly classified into end-to-end methods and pre-trained embedding methods.

End-to-end methods are tailored for specific tasks and are typically trained with task-specific labels. Trajectory prediction methods, such as DeepMove~\cite{DBLP:conf/www/FengLZSMGJ18}, HST-LSTM~\cite{DBLP:conf/ijcai/Kong018}, and ACN~\cite{DBLP:conf/wsdm/MiaoLZW20}, use Recurrent Neural Networks (RNN)~\cite{hochreiter1997long, DBLP:journals/corr/ChungGCB14} to capture sequential correlations in trajectories, while PreCLN~\cite{DBLP:journals/www/YanZSYD23} uses Transformers~\cite{DBLP:conf/nips/VaswaniSPUJGKP17} to process vehicle trajectories. Origin-destination travel time estimation methods, including TEMP~\cite{DBLP:conf/gis/WangKKL16}, MURAT~\cite{DBLP:conf/kdd/LiFWSYL18}, DeepOD~\cite{DBLP:conf/sigmod/Yuan0BF20}, and DOT~\cite{DBLP:journals/pacmmod/LinWHGYLJ23}, estimate travel time based on origin, destination, and departure time. Trajectory travel time estimation methods, including WDR~\cite{DBLP:conf/kdd/WangFY18}, DeepTTE~\cite{DBLP:conf/aaai/WangZCLZ18}, and WDDRA~\cite{DBLP:conf/gis/GanZW21}, predict travel time given the full trajectory. While end-to-end methods are straightforward to implement, they are not easily transferable to other tasks, requiring separate models for each task, which can strain computational and storage resources.

Recent efforts focus on building task-transferable methods primarily through pre-trained embedding techniques. Methods such as trajectory2vec~\cite{DBLP:conf/ijcnn/YaoZZHB17}, t2vec~\cite{DBLP:conf/icde/LiZCJW18}, Trembr~\cite{DBLP:journals/tist/FuL20}, START~\cite{DBLP:conf/icde/JiangPRJLW23}, and MMTEC~\cite{10375102} learn trajectory encoders that map vehicle trajectories into embedding vectors using pre-training techniques like auto-encoding~\cite{hinton2006reducing} and contrastive learning~\cite{DBLP:conf/icml/ChenK0H20}. Although these trajectory encoders show versatility, they require additional prediction modules to generate task-specific predictions from the embedding vectors. Additionally, their adaptability is limited for tasks involving incomplete trajectories.

\subsection{Foundation Models}
Foundation models refer to large, pre-trained models that serve as a base for various downstream tasks. These models excel in domains like Natural Language Processing (NLP) and Computer Vision (CV) due to the universal nature of languages and images, and the abundant data available in these fields.

Notable foundation models include BERT~\cite{DBLP:conf/naacl/DevlinCLT19}, GPT~\cite{DBLP:conf/naacl/DevlinCLT19}, and GLM~\cite{DBLP:conf/acl/DuQLDQY022} in NLP, and CLIP~\cite{DBLP:conf/icml/RadfordKHRGASAM21} in CV. BERT is used for tasks like question answering, sentiment analysis, and named entity recognition. GPT and GLM support tasks such as text generation, translation, and summarization. CLIP connects vision and language understanding, enabling tasks like image classification and captioning. However, these models are not directly adaptable for building a vehicle trajectory foundation model due to the unique challenges in dataset and task transferability for vehicle trajectories discussed in Section~\ref{sec:introduction}.

\section{Preliminaries} 
\label{sec:preliminaries}

\subsection{Definitions}
\begin{definition}
[Vehicle Trajectory]
A vehicle trajectory $\mathcal{T}$ is a sequence of trajectory points: $\mathcal{T} = \langle p_1, p_2, \dots, p_n \rangle$, where $n$ is the number of points. Each point $p_i = (\mathrm{lng}_i, \mathrm{lat}_i, t_i)$ consists of the longitude $\mathrm{lng}_i$, latitude $\mathrm{lat}_i$, and timestamp $t_i$, representing the vehicle's location at a specific time.
\end{definition}

\begin{definition}
[Trajectory Dataset]
A trajectory dataset, denoted as $\mathbb{T}$, is a collection of vehicle trajectories recorded in a specific region over a certain period. For each trajectory $\mathcal{T} \in \mathbb{T}$, the coordinates of its points fall within the region’s coordinate range, and the timestamps of its points fall within the designated time frame.
\end{definition}

\begin{definition}
[Point of Interest]
A point of interest (POI) is a significant geographical location with specific cultural, environmental, or economic importance. We represent a POI as $l_i = (\mathrm{lng}_i, \mathrm{lat}_i, \mathrm{desc}_i)$, where $\mathrm{lng}_i$ and $\mathrm{lat}_i$ denote the coordinates of the POI, and $\mathrm{desc}_i$ is a textual description including the name, type, and address of the POI.
\end{definition}

\subsection{Problem Statement}
\textbf{Region and task transferable vehicle trajectory learning} aims to develop a trajectory learning model $f_\theta$ with learnable parameters $\theta$. Once pre-trained, this model should effectively transfer between trajectory datasets covering different regions and accurately generate the required outputs for various tasks based on their inputs, all without needing to retrain the parameters $\theta$.

\begin{figure*}
    \centering
    \includegraphics[width=1.0\linewidth]{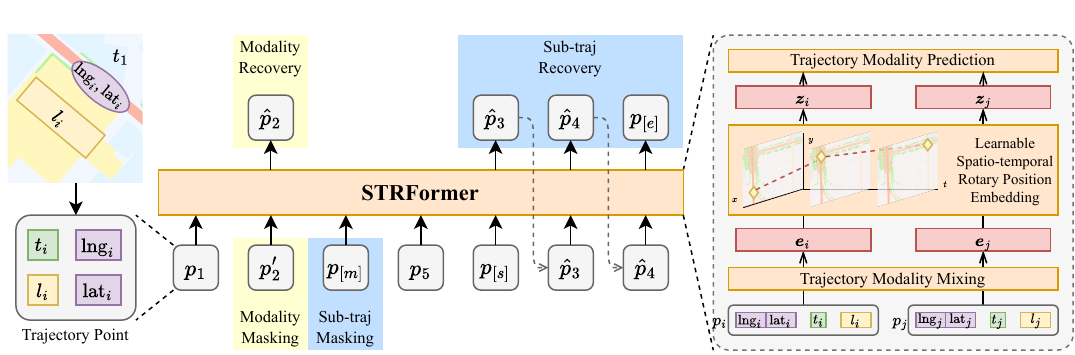}
    \caption{The framework of TrajFM.}
    \label{fig:framework}
\end{figure*}

\section{Methodology}
\subsection{Overview}
In this paper, we introduce TrajFM, a vehicle trajectory foundation model that excels in region and task transferability. The framework of TrajFM is illustrated in Figure~\ref{fig:framework}. It comprises two key components: the STRFormer, the core learnable model enabling region transferability; and the trajectory masking and recovery scheme, designed to enhance TrajFM's adaptability across diverse tasks.

The STRFormer is the learnable model of TrajFM. It initially maps each trajectory point into three feature modalities: spatial, temporal, and point of interest (POI). The inclusion of the POI modality allows the model to discern variations in POI arrangements across different regions. A modality mixing module then integrates these three modalities into a latent vector representing the trajectory point. Following this, a \textit{spatio-temporal Rotary position embedding} (\textbf{STRPE}) module, inspired by RoFormer~\cite{DBLP:journals/ijon/SuALPBL24}, captures the relative correlations between trajectory points, offering a transferable method to handle the diverse spatial features of different regions. These designs empower STRFormer with regional transferability.

The trajectory masking and recovery scheme equips TrajFM with the flexibility to handle various tasks. It standardizes different tasks by combining two approaches: masking and recovering modalities or sub-trajectories. In one approach, spatial or temporal modalities of a trajectory point are masked in the input and then recovered in the output. In another approach, a sub-trajectory is masked as a special mask point in the input and recovered in an auto-regressive manner in the output. Using this scheme during TrajFM's pre-training allows it to transfer between tasks seamlessly without re-training.

The subsequent sections provide an in-depth explanation of the components in TrajFM.

\subsection{STRFormer}
To enable region transferability, we propose a novel vehicle trajectory encoder called STRFormer. It first maps each point in a vehicle trajectory into spatial, temporal, and POI modalities. Next, a trajectory modality mixing module aggregates these modalities into an embedding vector for each point. Then, it models correlations between trajectory points using a learnable spatio-temporal Rotary position embedding (STRPE) module. Finally, a trajectory modality prediction module produces the final output.

\subsubsection{Trajectory Modalities}
We begin by representing each point $p_i$ in a trajectory $\mathcal{T}$ as a tuple of three modalities: spatial, temporal, and POI, denoted as $p_i=((\mathrm{lng}_i,\mathrm{lat}_i), t_i, l_i)$. The POI modality $l_i$ is obtained by locating the nearest POI to the trajectory point. Each modality is then processed and mapped into its latent embedding as detailed below.

For the spatial modality, we use the \textit{Universal Transverse Mercator} (UTM) projection to better align with the Earth's curvature. The projected coordinates are then normalized based on the center point of the area of interest. Formally, given $(\mathrm{lng}_i, \mathrm{lat}_i)$ and the center coordinates $(\mathrm{lng}_{\text{cen}}, \mathrm{lat}_{\text{cen}})$ of the trajectory dataset's area of interest, this process is formulated as follows:
\begin{equation}
    \begin{split}
    x_i &= (\mathrm{UTM}(\mathrm{lng}_i) - \mathrm{UTM}(\mathrm{lng}_\mathrm{cen}))/s_x\\
    y_i &= (\mathrm{UTM}(\mathrm{lat}_i) - \mathrm{UTM}(\mathrm{lat}_\mathrm{cen}))/s_y
 \end{split}
\label{eq:utm-projection}
\end{equation}
where $s_x$ and $s_y$ are scale factors shared across different datasets, which we set as $s_x=s_y=4000$, and $\mathrm{UTM}$ denotes the UTM projection function. Next, we use linear projection to obtain the embedding vector $\textbf{e}_i^s \in \mathbb{R}^d$ for the spatial modality, denoted as $\textbf{e}_i^s = \mathrm{Linear}(\langle x_i, y_i \rangle)$. Here, $\mathrm{Linear}$ is a fully-connected linear projection layer, and $d$ is the embedding dimension. This approach preserves the scale variations of different regions better than simple methods like min-max normalization.

For the temporal modality, we represent the timestamp $t_i$ using a vector $\boldsymbol{t}_i \in \mathbb{R}^4$ of four features to capture the periodicity of time: the day of the week, the hour of the day, the minute of the hour, and the time difference in minutes relative to $t_1$. These features are encoded into four embedding vectors using learnable Fourier encoding layers~\cite{DBLP:conf/nips/TancikSMFRSRBN20}. The embedding vectors are concatenated and mapped into the embedding vector $\textbf{e}_i^t \in \mathbb{R}^d$ of temporal modality through a linear projection layer.

For the POI modality, we input the textual description $\mathrm{desc}_i$ of $l_i$ into a pre-trained text embedding model, specifically the \textit{text-embedding-3-large} model from OpenAI\footnote{\url{https://platform.openai.com/docs/guides/embeddings}}, to derive a fixed-dimension embedding vector. This vector is then transformed into the embedding vector $\boldsymbol{e}_i^p \in \mathbb{R}^d$ of POI modality with a linear projection layer. This representation captures transferable semantic information and functionality of POIs, while also considering the diversity of POI arrangements across different regions.

\subsubsection{Trajectory Modality Mixing}
To capture the correlation between the three modalities of a trajectory point, we propose merging them into a single latent vector.

First, for added flexibility, each modality in a point $p_i$ can be replaced by special tokens: the mask token $[m]$, the start token $[s]$, or the end token $[e]$. If a modality is replaced by one of these tokens, the embedding vector for that modality, i.e., $\boldsymbol{e}_i^s$, $\boldsymbol{e}_i^t$, or $\boldsymbol{e}_i^p$, is substituted with the embedding vector of the corresponding token, denoted as $\boldsymbol{e}_{[m]}$, $\boldsymbol{e}_{[s]}$, and $\boldsymbol{e}_{[e]}$.

Next, we use a Transformer~\cite{DBLP:conf/nips/VaswaniSPUJGKP17} to mix and merge the latent vectors of the modalities. Specifically, we treat the latent vectors as a sequence of length 3 and compute the embedding vector $\boldsymbol{e}_i \in \mathbb{R}^d$ for point $p_i$ as follows:
\begin{equation}
    \boldsymbol{e}_i = \mathrm{MeanPool}(\mathrm{TransEnc}(\langle \boldsymbol{e}_i^s, \boldsymbol{e}_i^t, \boldsymbol{e}_i^p \rangle)),
\end{equation}
where $\mathrm{MeanPool}$ is a mean pooling layer, and $\mathrm{TransEnc}$ is a 1-layer Transformer encoder.

\subsubsection{Learnable Spatio-Temporal Rotary Position Embedding (STRPE)}
STRPE is designed to model the correlations between trajectory points' embedding vectors and relative spatial information simultaneously. Extracting this relative information helps prevent the model from becoming biased towards specific regions, providing a region transferable approach to handle spatial features.

STRPE is composed of $L$ layers. Each layer models the correlation between points in the input sequence using a specially designed attention mechanism. The output from the $(l-1)$-th layer serves as the input to the $l$-th layer. The input to the first layer is the sequence of embedding vectors for all points in a trajectory, formulated as $\langle \boldsymbol e_1, \boldsymbol e_2, \dots, \boldsymbol e_n \rangle$, where $n$ is the number of trajectory points.

To determine the relationship between the $i$-th and $j$-th trajectory points, we first introduce a learnable spatio-temporal Rotary matrix to encode spatial information. For a trajectory point with spatial modality $(x, y)$, the matrix is calculated as follows:
\begin{equation}
\begin{split}
    \mathbf{R}_{\Phi (x, y)} &= 
    \scalebox{0.65}{$
    \begin{bmatrix}  
        \text{cos}  \, \phi_1(x,y)\theta_1 & -\text{sin}  \, \phi_1(x,y)\theta_1 & \cdots & 0 & 0 \\  
        \text{sin} \, \phi_1(x,y)\theta_1 & \text{cos} \, \phi_1(x,y)\theta_1 & \cdots & 0 & 0 \\  
        \vdots & \vdots & \ddots & \vdots & \vdots \\  
        0 & 0 & \cdots & \text{cos} \, \phi_{d/2}(x,y)\theta_{d/2} & -\text{sin} \, \phi_{d/2}(x,y)\theta_{d/2} \\
        0 & 0 & \cdots & \text{sin} \, \phi_{d/2}(x,y)\theta_{d/2} & \text{cos}\, \phi_{d/2}(x,y)\theta_{d/2} 
    \end{bmatrix}$}
    \\
    \Phi (x, y) &= \boldsymbol W_\Phi(x||y),
\end{split}
\end{equation}
where $\theta_1, \theta_2, \dots, \theta_{d/2}$ are frequency weights and $\theta_k = 10000^{-2k/d}$. $\boldsymbol W_\Phi \in \mathbb{R}^{2 \times 2/d}$ is a mapping matrix, $\Phi (x, y)$ is a vector, and $\phi_k(x,y)$ is the $k$-th element of $\Phi (x,y)$.

Then, to model the correlation between embeddings $e_i$ and $e_j$, the query $\boldsymbol{q}_i$, key $\boldsymbol{k}_j$, and value $\boldsymbol{v}_j$ for the attention mechanism in this layer are calculated as follows:
\begin{equation}
    \begin{split}
        \boldsymbol{q}_i &= \mathbf{R}_{\Phi (x_i, y_i)} \boldsymbol W_q \boldsymbol{e}_i\\
        \boldsymbol{k}_j &= \mathbf{R}_{\Phi (x_j, y_j)} \boldsymbol W_k \boldsymbol{e}_j\\
        \boldsymbol{v}_j &= \boldsymbol W_v \boldsymbol{e}_j,
    \end{split}
\end{equation}
where $\boldsymbol W_q \in \mathbb{R}^{d \times d}$, $\boldsymbol W_k \in \mathbb{R}^{d \times d}$, and $\boldsymbol W_v \in \mathbb{R}^{d \times d}$ are mapping matrices. Following this, we implement the attention mechanism for the input sequence and calculate the $i$-th hidden state of this layer as follows:
\begin{equation}
    \boldsymbol h_i = \sum_{j=1}^{n} \frac{\exp(\frac{{\boldsymbol{q}_i}^\top \cdot \boldsymbol{k}_j }{\sqrt{d} } ) }{ {\textstyle \sum_{j=1}^{n}} \exp(\frac{{\boldsymbol{q}_i}^\top \cdot \boldsymbol{k}_j }{\sqrt{d} } )} \boldsymbol{v}_j 
\end{equation}

Since the attention mechanism involves modeling the correlation between trajectory points with dot-product, the proposed learnable spatio-temporal Rotary matrix enhances the model's ability to capture relative spatial information between trajectory points. This is shown in the expanded formula of the dot-product:
\begin{equation}
\begin{split}
    {\boldsymbol{q}_i}^\top \cdot \boldsymbol{k}_j &= (\mathbf{R}_{\Phi (x_i, y_i)} \boldsymbol W_q  {\boldsymbol{e}_i})^\top \cdot (\mathbf{R}_{\Phi (x_j, y_j)} \boldsymbol W_k  \boldsymbol{e}_j) \\
    &= {\boldsymbol{e}_i}^\top {\boldsymbol W_q }^\top \mathbf{R}_{\Phi (x_i, y_i)}^\top (\mathbf{R}_{\Phi (x_j, y_j)} \boldsymbol W_k  \boldsymbol{e}_j) \\
    &= {\boldsymbol{e}_i}^\top {\boldsymbol W_q}^\top \mathbf{R}_{\Phi (x_i, y_i)-\Phi (x_j, y_j)} \boldsymbol W_k \boldsymbol{e}_j,
\end{split}
\end{equation}
where $\mathbf{R}_{\Phi (x_i, y_i) - \Phi (x_j, y_j)}$ represents learnable relative spatial information extracted from the differences in spatial features between the $i$-th and $j$-th trajectory points. This characteristic ensures that STRPE can handle differences in spatial features between regions in a transferable way.

Finally, we implement layer normalizations and a feed-forward network to obtain the output of this layer. The $i$-th output of this layer is calculated as follows:
\begin{equation}
    \boldsymbol{h}_i' = \text{LayerNorm}(\text{FFN}(\text{LayerNorm}(\boldsymbol{h}_i + \boldsymbol{e}_i)) + \boldsymbol{h}_i),
\end{equation}
which forms the output sequence of this layer and also serves as the input to the next layer.

The process described for the first layer of STRPE can be generalized to other layers. Note that different layers do not share learnable parameters. The $i$-th output from the last layer of STRPE is regarded as the latent vector of the $i$-th trajectory point, denoted as $\boldsymbol z_i$.

\subsubsection{Trajectory Modality Prediction}
After STRPE derives the latent vector $\boldsymbol{z}_i$ for each trajectory point $p_i$, we use prediction modules on the vector to predict the modalities in $p_i$.

For the spatial modality, we use a linear projection layer with an output dimension of 2 to predict its UTM-projected and normalized coordinates, i.e., $\hat{x}_i$ and $\hat{y}_i$. We can then obtain the coordinates $(\hat{\mathrm{lng}}_i, \hat{\mathrm{lat}}_i)$ by reversing the process in Equation~\ref{eq:utm-projection}. The POI modality is then recovered by finding the nearest POI to the predicted coordinates. For the temporal modality, we use a linear projection layer with an output dimension of 4 followed by Softplus activation to predict the expanded temporal features, i.e., $\hat{\boldsymbol{t}}_i$.

To supervise the predicted modalities, we apply the Mean Squared Error (MSE) loss function to the predicted spatial and temporal modalities, formulated as follows:
\begin{equation}
\begin{split}
    \mathcal{L}_i^s &= (\hat{x}_i - x_i)^2 + (\hat{y}_i - y_i)^2, \\
    \mathcal{L}_i^t &= \Vert \hat{\boldsymbol{t}}_i - \boldsymbol{t}_i \Vert_2
\end{split}
\label{eq:spatial-temporal-loss}
\end{equation}

Additionally, we implement a token predictor for each modality to enable the prediction of a special end token $[e]$. The token predictor is a linear projection layer with a Sigmoid activation, outputting a probability $\hat{r}$ of whether the modality is $[e]$. If $\hat{r} \geq 0.5$, then the corresponding predicted modality is set to $[e]$. We supervise this prediction using cross-entropy loss:
\begin{equation}
    \mathcal{L}_i^e = -(r\log(\hat{r}) + (1 - r)\log(1 - \hat{r})),
\label{eq:token-loss}
\end{equation}
where $r = 1$ if the ground-truth modality is $[e]$ and $r = 0$ otherwise.

\subsection{Trajectory Masking and Recovery Scheme}
To enhance task transferability, we present a new trajectory masking and recovery scheme that unifies generation processes of various tasks. It combines two approaches: 1) masking a modality of a single trajectory point from the input and recovering it in the output, and 2) masking a sub-trajectory from the input using a special mask point and auto-regressively recovering it in the output. TrajFM is pre-trained using this scheme by randomly applying the two approaches.

\subsubsection{Modality Masking and Recovery}
\label{sec:modality-masking-and-recovery}
Given a trajectory point $p_i=((\mathrm{lng}_i,\mathrm{lat}_i), t_i, l_i)$ in its three modalities, we can mask either its spatial or temporal modalities. A point with a masked spatial modality is represented as $p^{ms}_i=([m], t_i, [m])$, with its spatial and POI modalities replaced by the special mask token $[m]$. Note that the POI modality is linked to the spatial modality, so masking the spatial modality will also mask the POI modality. Similarly, a point with a masked temporal modality is denoted as $p^{mt}_i=((\mathrm{lng}_i,\mathrm{lat}_i), [m], l_i)$.

The masked modality in $p^{ms}_i$ or $p^{mt}_i$ is then recovered on the corresponding output of the trajectory modality prediction module at the same time step, i.e., $\hat p_i$.

\subsubsection{Sub-trajectory Masking and Recovery}
\label{sec:sub-trajectory-masking-and-recovery}
Given a trajectory $\mathcal T=\langle p_1, p_2, \dots, p_n \rangle$, we can mask a sub-trajectory $\langle p_s, p_{s+1}, \dots, p_{e} \rangle$ from it by replacing the sub-trajectory with a special mask point $p_{[m]}=([m],[m],[m])$, where all three modalities are the special mask token. The masked trajectory is denoted as $\mathcal T^m=\langle p_1, \dots, p_{s-1}, p_{[m]}, \\p_{e+1}, \dots, p_n \rangle$.

The masked sub-trajectory is then recovered auto-regressively by extending the input and output sequence of STRFormer. After feeding $\mathcal T^m$ as the input, a special start point $p_{[s]}=([s],[s],[s])$ with modalities being the special start token $[s]$ is appended to the input sequence. The trajectory modality prediction module produces the recovered first point of the sub-trajectory $\hat p_s$ at the last step of the output sequence. This recovered point is then appended to the input sequence, and the procedure is repeated until the entire sub-trajectory is recovered. The recovery process ends by appending the recovered last point of the sub-trajectory $\hat p_e$ to the input, and the trajectory modality prediction module produces a special end point $p_{[e]}=([e],[e],[e])$ with modalities being the special end token $[e]$.

\subsubsection{Pre-training with the Scheme}
We propose to pre-train TrajFM with a mixture of the above two masking and recovery approaches, enabling it to transfer between task transferability without re-training.

Given a trajectory $\mathcal T$ in the pre-train dataset, we first randomly select the starting $s$ and ending $e$ points for the sub-trajectory masking from a uniform distribution $U(1, n)$, with $s < e$. The selected sub-trajectory, $\langle p_s, p_{s+1}, \dots, p_{e} \rangle$, is then masked from the input and recovered in the output following the process detailed in Section~\ref{sec:sub-trajectory-masking-and-recovery}.
For the remaining trajectory points in the input, $p_1, \dots, \\p_{s-1}, p_{e+1}, \dots, p_n$, we randomly mask their spatial or temporal modalities with a even probability, and recover the masked modalities in the output following the procedure in Section~\ref{sec:modality-masking-and-recovery}.
Finally, the pre-training loss of $\mathcal T$ can be derived by adding the losses of all recovered modalities and predicted tokens, with each loss of one modality or token detailed in Equations~\ref{eq:spatial-temporal-loss} and \ref{eq:token-loss}.

\subsubsection{Transfer to Tasks}
\label{sec:transfer-to-tasks}
By combining the masking and recovery scheme that unifies diverse generation schemes for different tasks with the pre-training procedure, TrajFM can transfer to various tasks without retraining. Here, we present three representative types of tasks: trajectory travel time estimation, origin-destination travel time estimation, and trajectory prediction.

The \textbf{trajectory travel time estimation} task aims to predict the travel time of a trajectory based on its spatial information and departure time. For this task, the trajectory with the temporal modality masked at each point except the first one serves as the input. The predicted travel time is derived from the recovered temporal modality of the last point $\hat p_n$, following the method in Section~\ref{sec:modality-masking-and-recovery}. Mean Absolute Error (MAE), Root Mean Squared Error (RMSE), and Mean Absolute Percentage Error (MAPE) of the travel time are used as evaluation metrics.

The \textbf{origin-destination travel time estimation} task aims to predict the travel time of a trajectory given its starting and ending locations and departure time. For this task, a sequence $\langle p_1, [m], p_n' \rangle$ can be used as the input to TrajFM, where $p_n'=((\mathrm{lng}_n, \mathrm{lat}_n), [m], l_n)$. Similar to the previous task, the predicted travel time is derived from the recovered temporal modality of the last point $\hat p_n$. MAE, RMSE, and MAPE are used as evaluation metrics.

The \textbf{trajectory prediction} task aims to forecast the future segment of a trajectory given its historical segment. For this task, the historical portion of the trajectory, ending with a mask point, can be provided to TrajFM as $\langle p_1, p_2, \dots, p_{n'}, [m]\rangle$, where $n'$ is the historical length. The future portion is then predicted using the procedure detailed in Section~\ref{sec:sub-trajectory-masking-and-recovery}. We set $n'=n-5$, and evaluate the precision of the trajectories' destinations. MAE and RMSE of the shortest distance on the Earth's surface are used as evaluation metrics.

\section{Experiments}
We assess the effectiveness of TrajFM using two real-world vehicle trajectory datasets under various experimental settings.

\subsection{Datasets}
The two vehicle trajectory datasets, referred to as \textbf{Chengdu} and \textbf{Xian}, consist of vehicle trajectories recorded by taxis operating in Chengdu and Xian, China, and were released by Didi\footnote{\url{https://gaia.didichuxing.com/}}. Due to the original trajectories having very dense sampling intervals, we retain a portion of the trajectory points using a three-hop resampling method, resulting in most trajectories having sampling intervals no shorter than 6 seconds. After resampling, trajectories with fewer than 5 or more than 120 trajectory points, considered as anomalies, are excluded.
Additionally, we retrieve information on POIs within these datasets' areas of interest from the AMap API\footnote{\url{https://lbs.amap.com/api/javascript-api-v2}}.
The statistics of these datasets after preprocessing are listed in Table~\ref{table:dataset}.

\begin{table}[h]
    \centering
    \caption{Dataset statistics.}
    \begin{tabular}{ccc}
    \toprule
    Dataset & Chengdu & Xian \\
    \midrule
    Time span & 09/30 - 10/10, 2018 & 09/29 - 10/15, 2018 \\
    \#Trajectories & 140,000 & 210,000 \\
    \#Points & 18,832,411 & 18,267,440 \\
    \#POIs & 12,439 & 3,900 \\
    \bottomrule
\end{tabular}

    \label{table:dataset}
\end{table}

\subsection{Comparison Methods}
Several state-of-the-art vehicle trajectory learning methods are included in the experiments for comparison.

\begin{itemize}[leftmargin=*]
    \item \textbf{t2vec}~\cite{DBLP:conf/kdd/Fang0ZHCGJ22}: Pre-trains the model by reconstructing original trajectories from low-sampling ones using a denoising auto-encoder.
    \item \textbf{Trembr}~\cite{DBLP:journals/tist/FuL20}: Constructs an RNN-based seq2seq model that recovers the road segments and time of the input trajectories.
    \item \textbf{CTLE}~\cite{DBLP:conf/aaai/LinW0L21}: Pre-trains a bi-directional Transformer with two MLM tasks involving location and hour predictions. The trajectory representation is obtained by applying mean pooling on point embeddings.
    \item \textbf{Toast}~\cite{DBLP:conf/cikm/ChenLCBLLCE21}: Utilizes a context-aware node2vec model to generate segment representations and trains the model with an MLM-based task and a sequence discrimination task.
    \item \textbf{TrajCL}~\cite{DBLP:conf/icde/Chang0LT23}: Introduces a dual-feature self-attention-based encoder and trains the model in a contrastive style using the InfoNCE loss.
    \item \textbf{START}~\cite{DBLP:conf/icde/JiangPRJLW23}: Includes a time-aware trajectory encoder and a GAT that considers the transitions between road segments. The model is trained with both an MLM task and a contrastive task based on SimCLR loss.
    \item \textbf{LightPath}~\cite{DBLP:conf/kdd/YangHGYJ23}: Constructs a sparse path encoder and trains it with a path reconstruction task and a cross-view and cross-network contrastive task.
\end{itemize}

Since these methods output embeddings for input trajectories, we apply these embeddings to fully connected networks to make predictions for downstream tasks. We can either 1) fine-tune the methods' parameters using task supervision, or 2) fix their parameters and only update the predictors' parameters. In the experiments, we denote the latter setting as \textit{without fine-tune} (\textbf{wo ft}).

\subsection{Settings}
For both datasets, we split the departure time of the trajectories chronologically into 8:1:1 ratios to create the training, validation, and testing sets. TrajFM is pre-trained for 30 epochs on the training set, and the final metrics of downstream tasks are calculated on the testing set with its parameters fixed.

The two key hyper-parameters of TrajFM and their optimal values are $L=2$ and $d=128$. A detailed analysis of their impact on performance is provided in  Appendix~\ref{sec:hyper-parameters}. We choose parameters based on the MAE and RMSE of the trajectory prediction task on Chengdu's validation set. We also analyze the effectiveness of modules in TrajFM in Appendix~\ref{sec:modules}. 
TrajFM is implemented using PyTorch~\cite{DBLP:conf/nips/PaszkeGMLBCKLGA19}. For model training, we utilize the Adam optimizer with an initial learning rate of 1e-3. The experiments are conducted on Ubuntu 22.04 servers equipped with Intel(R) Xeon(R) W-2155 CPUs and nVidia(R) TITAN RTX GPUs. We run each set of experiments 5 times and report the mean and deviation of the metrics.

\begin{table*}
    \caption{Overall performance of methods on trajectory arrival time estimation.}
    \label{tab:overall-trajectory-eta}
\begin{threeparttable}
\begin{tabular}{c|ccc|ccc}
\toprule
Dataset & \multicolumn{3}{c|}{Chengdu} & \multicolumn{3}{c}{Xian} \\
\midrule
\multirow{2}{*}{\diagbox[]{Method}{Metric}} & RMSE $\downarrow$ & MAE $\downarrow$ & MAPE $\downarrow$ & RMSE $\downarrow$ & MAE $\downarrow$ & MAPE $\downarrow$ \\
& (seconds) & (seconds) & (\%) & (seconds) & (seconds) & (\%) \\
\midrule
Trembr (wo ft)  & 179.66$\pm$0.49 &  127.40$\pm$0.20 & 36.50$\pm$0.41 & 368.34$\pm$3.14 & 277.85$\pm$2.77 & 42.93$\pm$6.57 \\
START (wo ft)  & 132.71$\pm$1.24 &  65.88$\pm$0.54 & 16.12$\pm$0.27 & 162.34$\pm$8.12 & 65.23$\pm$2.96 & 8.06$\pm$0.17 \\
LightPath (wo ft)  & 134.95$\pm$1.59 &  73.44$\pm$2.13 & 18.95$\pm$1.34 & 194.83$\pm$1.67 & 97.76$\pm$3.04 & 15.73$\pm$1.25 \\
\midrule
t2vec  & 128.51$\pm$2.60 & 60.52$\pm$2.58 & 15.22$\pm$0.45 & 199.13$\pm$2.45 & 86.01$\pm$2.83 & 14.22$\pm$0.50 \\
Trembr  & 125.54$\pm$2.85 & 57.97$\pm$2.59 & 13.96$\pm$0.86 & 185.73$\pm$3.56 & 81.12$\pm$2.41 & 12.77$\pm$0.77 \\
CTLE  & 132.64$\pm$3.97 & 57.48$\pm$1.14 & 13.15$\pm$0.75 & 182.28$\pm$2.67 & \underline{79.71$\pm$1.62} & 12.78$\pm$0.57 \\
Toast  & 128.79$\pm$2.57 & 61.00$\pm$3.54 & 14.88$\pm$0.58 & 183.09$\pm$3.83 & 84.93$\pm$2.47 & 13.44$\pm$0.63 \\
TrajCL  & 120.21$\pm$1.04 & 59.82$\pm$1.84 & 14.74$\pm$0.44 & \underline{179.81$\pm$3.30} & 82.49$\pm$2.91 & 13.23$\pm$0.27 \\
START  & 122.21$\pm$3.18 & 55.92$\pm$2.40 & 12.72$\pm$0.79 & 182.35$\pm$3.25 & 80.76$\pm$2.76 & 12.55$\pm$0.50 \\
LightPath  & \underline{119.23$\pm$2.37} & \underline{55.61$\pm$1.52} & \underline{12.76$\pm$0.85} & 180.03$\pm$2.37 & 80.42$\pm$2.19 & \underline{12.25$\pm$0.69}\\

\textbf{TrajFM (ours)} & \textbf{113.55$\pm$1.88} & \textbf{50.63$\pm$1.98} & \textbf{11.02$\pm$0.97} & \textbf{170.49$\pm$2.24} & \textbf{75.37$\pm$1.87} & \textbf{10.61$\pm$1.00} \\
\bottomrule
\end{tabular}
\begin{tablenotes}\footnotesize
\item[]{
    \textbf{Bold} denotes the best result, and \underline{underline} denotes the second-best result.
    $\downarrow$ means lower is better.
}
\end{tablenotes}
\end{threeparttable}
\end{table*}

\begin{table*}
    \caption{Overall performance of methods on origin-destination arrival time estimation.}
    \label{tab:overall-od-eta}
\begin{threeparttable}
\begin{tabular}{c|ccc|ccc}
\toprule
Dataset & \multicolumn{3}{c|}{Chengdu} & \multicolumn{3}{c}{Xian} \\
\midrule
\multirow{2}{*}{\diagbox[]{Method}{Metric}} & RMSE $\downarrow$ & MAE $\downarrow$ & MAPE $\downarrow$ & RMSE $\downarrow$ & MAE $\downarrow$ & MAPE $\downarrow$ \\
& (seconds) & (seconds) & (\%) & (seconds) & (seconds) & (\%) \\

\midrule

Trembr (wo ft)  & 337.07$\pm$2.37 & 239.97$\pm$1.68 & 31.40$\pm$0.09 & 276.21$\pm$3.03 & 218.69$\pm$3.43 & 53.09$\pm$0.56\\

START (wo ft)  & 243.28$\pm$5.17
& 186.61$\pm$4.43
& 49.42$\pm$0.41
& 392.80$\pm$1.28
& 313.19$\pm$1.31
& 71.42$\pm$0.46 \\

LightPath (wo ft)  & 232.02$\pm$3.40 &  180.92$\pm$2.61 & 52.91$\pm$0.93 & 397.78$\pm$4.63 & 318.54 $\pm$ 4.43 & 70.92 $\pm$ 0.83 \\

\midrule

t2vec  & 235.99$\pm$4.55
& 182.26$\pm$3.91
& 49.37$\pm$0.40
& 390.54$\pm$0.68
& 311.29$\pm$0.13
& 71.46$\pm$0.22 \\

Trembr   & 232.44$\pm$1.90
& 166.29$\pm$2.55
& 40.59 $\pm$0.50
& 360.67$\pm$4.39
& 281.28$\pm$6.30
& 58.62$\pm$1.04\\

CTLE   & 232.26$\pm$15.55
& 157.15$\pm$10.85
& 39.00$\pm$3.20
& 240.99$\pm$15.78
& 134.20$\pm$8.98
& 22.58$\pm$0.71\\

Toast   & 222.42$\pm$13.16
& \underline{150.81$\pm$8.61}
& \underline{36.90 $\pm$7.25}
& 260.21$\pm$19.10
& 165.92$\pm$4.55
& 35.69$\pm$2.01 \\

TrajCL   & 229.29$\pm$16.60
& 176.55$\pm$16.06
& 50.30 $\pm$1.97
& 365.02$\pm$12.69
& 283.18$\pm$13.55
& 61.20$\pm$4.60\\

START   & 225.46$\pm$13.46
& 170.95$\pm$14.66
& 47.17$\pm$3.44
& \underline{203.55$\pm$12.78}
& \underline{110.36$\pm$3.65}
& \underline{20.91$\pm$0.32}\\

LightPath  & \underline{219.47$\pm$20.48}
& 160.36$\pm$18.49
& 44.44$\pm$2.30
& 348.14$\pm$20.92
& 256.51$\pm$14.42
& 44.45$\pm$2.23 \\

\textbf{TrajFM (ours)} & \textbf{94.28$\pm$6.10} & \textbf{47.96$\pm$4.81} & \textbf{11.09$\pm$1.09} & \textbf{193.61$\pm$6.44} & \textbf{102.02$\pm$0.70} & \textbf{12.92$\pm$1.11} \\
\bottomrule
\end{tabular}
\begin{tablenotes}\footnotesize
\item[]{
    \textbf{Bold} denotes the best result, and \underline{underline} denotes the second-best result.
    $\downarrow$ means lower is better.
}
\end{tablenotes}
\end{threeparttable}
\end{table*}

\begin{table}
    \caption{Overall performance of methods on trajectory prediction.}
    \label{tab:overall-trajectory-prediction}
    \resizebox{1.0\linewidth}{!}{
\begin{threeparttable}
\begin{tabular}{c|cc|cc}
\toprule
Dataset & \multicolumn{2}{c|}{Chengdu} & \multicolumn{2}{c}{Xian} \\
\midrule
\multirow{2}{*}{\small \diagbox[]{Method}{Metric}} & RMSE $\downarrow$ & MAE $\downarrow$ & RMSE $\downarrow$ & MAE $\downarrow$ \\
& (meters) & (meters) & (meters) & (meters) \\
\midrule
Trembr (wo ft) & 1787.18$\pm$29.01 & 1419.58$\pm$28.95 & 2067.80$\pm$16.30 & 1749.76$\pm$18.82\\
START (wo ft) & 1347.13$\pm$30.72 & 1111.77$\pm$29.11 & 1406.06$\pm$18.42 & 1173.62$\pm$17.18\\
LightPath (wo ft)  & 2365.87$\pm$57.52 &  1948.97$\pm$57.78 & 2177.27$\pm$60.03 & 1859.35$\pm$48.50 \\
\midrule
t2vec &579.30$\pm$11.94 & 387.50$\pm$4.03 & 482.64$\pm$2.67 & 310.08$\pm$3.00\\
Trembr & 505.62$\pm$4.57 & 376.88$\pm$7.34 & 473.97$\pm$1.24 & 301.45$\pm$4.98\\
CTLE & 430.19$\pm$52.64 & 382.82$\pm$52.88 & 477.70$\pm$48.25 & 384.08$\pm$53.18\\
Toast  & 480.52$\pm$82.39 & 412.58$\pm$72.32 & 523.76$\pm$67.04 & 443.99$\pm$60.41\\
TrajCL & 365.50$\pm$19.14 & 272.63$\pm$25.32 & 383.39$\pm$7.30 & 262.20$\pm$10.68\\
START & \underline{333.10$\pm$10.47} & \underline{240.40$\pm$15.10} & \underline{319.00$\pm$4.27} & \underline{208.35$\pm$7.30}\\
LightPath & 553.27$\pm$42.26 & 360.86$\pm$56.41 & 598.20$\pm$15.57 & 348.61$\pm$19.32\\
\textbf{TrajFM (ours)} & \textbf{196.10$\pm$20.73} & \textbf{150.86$\pm$23.31} & \textbf{320.32$\pm$9.81} &  \textbf{207.89$\pm$5.61} \\
\bottomrule
\end{tabular}
\begin{tablenotes}\footnotesize
\item[]{
    \textbf{Bold} denotes the best result, and \underline{underline} denotes the second-best result.
    $\downarrow$ means lower is better.
}
\end{tablenotes}
\end{threeparttable}
}
\end{table}

\subsection{Performance Comparison}
\subsubsection{Overall Performance}
\label{sec:overall-performance}
Tables~\ref{tab:overall-trajectory-eta} to~\ref{tab:overall-trajectory-prediction} compare the overall performance of different methods on the tasks introduced in Section~\ref{sec:transfer-to-tasks}. Additionally, we present the full results of the performance of comparison methods under the \textit{wo ft} setting in Appendix~\ref{sec:wo-ft-variants}. TrajFM consistently shows promising performance, especially considering that, unlike comparison methods, it does not need any additional training besides pre-training to achieve this performance.

We observe that most comparison methods’ \textit{wo ft} setting have significant performance degradation compared to the fully fine-tuned approach. To reach optimal performance, these methods require fine-tuning the entire trajectory learning model with task supervision. This means that in real-world scenarios where different tasks are performed, multiple instances of these methods need to be re-trained and stored, failing to fully achieve task transferability. By comparison, TrajFM does not require fine-tuning the learnable model or prediction modules to reach the reported performance. It can be pre-trained once and directly perform different tasks with high performance. This demonstrates its superior task transferability, resulting in high efficiency and utilization in real-world applications.

We also observe that the performance gap between comparison methods and TrajFM is particularly evident in origin-destination arrival time estimation and trajectory prediction tasks. These tasks involve incomplete trajectory input, posing a great challenge for comparison methods due to their reliance on input integrity. In contrast, TrajFM handles these tasks effectively, thanks to its ability to adapt to incomplete input.

\begin{table*}
    \caption{Region transfer performance of methods on trajectory arrival time estimation.}
    \label{tab:transfer-trajectory-eta}
    \begin{threeparttable}
\begin{tabular}{c|ccc|ccc}
\toprule
Dataset & \multicolumn{3}{c|}{Chengdu $\rightarrow$ Xian} & \multicolumn{3}{c}{Xian $\rightarrow$ Chengdu} \\
\midrule
\multirow{2}{*}{\diagbox[]{Method}{Metric}} & RMSE $\downarrow$ & MAE $\downarrow$ & MAPE $\downarrow$ & RMSE $\downarrow$ & MAE $\downarrow$ & MAPE $\downarrow$ \\
& (seconds) & (seconds) & (\%) & (seconds) & (seconds) & (\%) \\
\midrule
Trembr (wo ft)  & 337.07$\pm$2.37 & 239.97$\pm$1.68 & 31.40$\pm$0.09 & 276.21$\pm$3.03 & 218.69$\pm$3.43 & 53.09$\pm$0.56\\
START (wo ft)  & \underline{177.40$\pm$3.29} & 70.81$\pm$3.14 & \underline{9.06$\pm$0.46} & 132.12$\pm$0.81 & 66.03$\pm$0.79 & 16.40$\pm$0.36\\
LightPath (wo ft)  & 181.20$\pm$7.80 & \underline{70.23$\pm$8.71} & 9.30$\pm$1.78 & \underline{119.93$\pm$6.07} & \underline{65.60$\pm$8.08} & \underline{13.32$\pm$2.97}\\
\textbf{TrajFM (ours)}  & \textbf{173.24$\pm$1.98} & \textbf{66.92$\pm$1.08} & \textbf{8.99$\pm$1.04} & \textbf{115.18$\pm$1.38} & \textbf{60.04$\pm$0.84} & \textbf{12.64$\pm$0.72}\\
\bottomrule
\end{tabular}
\begin{tablenotes}\footnotesize
\item[]{
    \textbf{Bold} denotes the best result, and \underline{underline} denotes the second-best result.
    $\downarrow$ means lower is better.
}
\end{tablenotes}
\end{threeparttable}
\end{table*}

\begin{table*}
    \caption{Region transfer performance of methods on origin-destination arrival time estimation.}
    \label{tab:transfer-od-eta}
    \begin{threeparttable}
\begin{tabular}{c|ccc|ccc}
\toprule
Dataset & \multicolumn{3}{c|}{Chengdu $\rightarrow$ Xian} & \multicolumn{3}{c}{Xian $\rightarrow$ Chengdu} \\
\midrule
\multirow{2}{*}{\diagbox[]{Method}{Metric}} & RMSE $\downarrow$ & MAE $\downarrow$ & MAPE $\downarrow$ & RMSE $\downarrow$ & MAE $\downarrow$ & MAPE $\downarrow$ \\
& (seconds) & (seconds) & (\%) & (seconds) & (seconds) & (\%) \\
\midrule

Trembr (wo ft)   & 352.13$\pm$3.92
& 274.28$\pm$4.36
& 55.33$\pm$2.25
& \underline{209.34$\pm$2.45}
& \underline{151.89$\pm$2.60}
& \underline{41.50$\pm$2.27}\\

START (wo ft)   & \underline{233.81$\pm$13.48}
& \underline{138.12$\pm$3.78}
& \underline{17.68$\pm$1.20}
& 235.61$\pm$3.19
& 182.60$\pm$2.35
& 51.20$\pm$0.79\\

LightPath (wo ft)   & 472.12$\pm$5.53
& 367.56$\pm$9.12
& 52.35$\pm$8.00 
& 231.08$\pm$1.17
& 168.40$\pm$2.10
& 42.87$\pm$1.95\\

\textbf{TrajFM (ours)} & \textbf{217.44$\pm$9.06} & \textbf{118.08$\pm$2.16} & \textbf{15.87$\pm$1.63} & \textbf{110.76$\pm$6.60} & \textbf{59.04$\pm$4.02} & \textbf{14.16$\pm$1.19} \\
\bottomrule
\end{tabular}
\begin{tablenotes}\footnotesize
\item[]{
    \textbf{Bold} denotes the best result, and \underline{underline} denotes the second-best result.
    $\downarrow$ means lower is better.
}
\end{tablenotes}
\end{threeparttable}
\end{table*}

\begin{table}
    \caption{Region transfer performance of methods on trajectory prediction.}
    \label{tab:transfer-trajectory-prediction}
    \resizebox{1.0\linewidth}{!}{
\begin{threeparttable}
\begin{tabular}{c|cc|cc}
\toprule
Dataset & \multicolumn{2}{c|}{Chengdu $\rightarrow$ Xian} & \multicolumn{2}{c}{Xian $\rightarrow$ Chengdu} \\
\midrule
\multirow{2}{*}{\small \diagbox[]{Method}{Metric}} & RMSE $\downarrow$ & MAE $\downarrow$ & RMSE $\downarrow$ & MAE $\downarrow$ \\
& (meters) & (meters) & (meters) & (meters) \\
\midrule
Trembr (wo ft) & 2914.69$\pm$16.00 & 2566.69$\pm$17.19 & 2706.51$\pm$93.94 & 2234.04$\pm$80.87\\
START (wo ft) & \underline{1891.36$\pm$14.28} & \underline{1627.67$\pm$14.58} & \underline{1859.68$\pm$25.19} & \underline{1569.15$\pm$19.42}\\
LightPath (wo ft) & 2793.72$\pm$27.90 & 2465.34$\pm$20.76 & 2610.30$\pm$66.11 & 2228.35$\pm$62.90 \\
\textbf{TrajFM (ours)} & \textbf{375.06$\pm$5.75} & \textbf{277.35$\pm$4.86} & \textbf{288.04$\pm$13.43} & \textbf{212.32$\pm$9.09} \\
\bottomrule
\end{tabular}
\begin{tablenotes}\footnotesize
\item[]{
    \textbf{Bold} denotes the best result, and \underline{underline} denotes the second-best result.
    $\downarrow$ means lower is better.
}
\end{tablenotes}
\end{threeparttable}
}
\end{table}

\subsubsection{Region Transfer Performance}
Tables~\ref{tab:transfer-trajectory-eta} to~\ref{tab:transfer-trajectory-prediction} compare the region transfer performance of methods. Both comparison methods and TrajFM are trained on one dataset and then directly evaluated on another dataset without further fine-tuning. TrajFM shows superior performance, demonstrating its region transferability.

As observed in the previous section, most comparison methods perform poorly without task supervision fine-tuning. The region transfer setting is more challenging since the methods operate in a different region from where they were trained and without task supervision. Comparatively, TrajFM is effective in region transfer. We also observe that in some cases, TrajFM performs better when transferred from another region compared to being directly pre-trained on the region. This shows that TrajFM can transfer from regions with higher quality training data to regions with poor data availability, highlighting the benefits of region transferability.

\begin{table}
    \caption{Efficiency of methods.}
    \label{tab:efficiency}
    \begin{threeparttable}
\begin{tabular}{c|ccc}
\toprule
Dataset & \multicolumn{3}{c}{Chengdu / Xian} \\
\midrule
\multirow{2}{*}{\small \diagbox[]{Method}{Metric}} & Model size & Train time & Test time \\
& (MBytes) & (min/epoch) & (seconds) \\
\midrule
t2vec & \textbf{1.64} / 6.30 & \underline{2.78} / \underline{5.94} & \underline{4.45} / \textbf{9.70}\\
Trembr & 5.75 / 5.30 & 3.36 / 6.07& \textbf{3.23} / \underline{9.72} \\
CTLE & 3.76 / 3.76 & 4.53 / 14.35 & 14.58 / 33.86\\
Toast & 4.01 / \underline{3.56} & 4.40 / 10.65 & 14.54 / 33.86 \\
TrajCL & 4.38 / 3.93 & 7.70 / 14.57 & 10.25 / 23.88 \\
START & 15.93 / 15.03 & 15.93 / 37.53 & 28.70 / 49.89 \\
LightPath & 12.96 / 12.51 & 10.25 / 23.22 & 22.49 / 46.26\\
\textbf{TrajFM (ours)} & \underline{3.27} / \textbf{3.27} & \textbf{0.94} / \textbf{3.03} & 15.42 / 30.94 \\
\bottomrule
\end{tabular}
\begin{tablenotes}\footnotesize
\item[]{
    \textbf{Bold} denotes the best result, and \underline{underline} denotes the second-best result.
}
\end{tablenotes}
\end{threeparttable}
\end{table}

\subsubsection{Efficiency}
Table~\ref{tab:efficiency} compares the efficiency of different methods on both datasets, measured by the size of the learning model and the time required for training and testing. We observe that TrajFM is lightweight, with a model size comparable to RNN-based methods like t2vec and Trembr, and much smaller than state-of-the-art methods like START and LightPath. TrajFM also has an efficient training process and is competitive at test time compared to other methods. Overall, TrajFM enhances the efficiency of real-world applications, as it only needs to be trained once and can perform various tasks in different regions without re-training.

\section{Conclusion}
We propose TrajFM, a vehicle trajectory foundation model that excels in region and task transferability. First, we introduce STRFormer as TrajFM's learnable model to enable region transferability. It integrates the POI modality of trajectories to adapt to variations in POI arrangements across regions. Additionally, it incorporates a learnable spatio-temporal Rotary position embedding module as a region-transferable approach for handling spatial features. Second, we propose a trajectory masking and recovery scheme, which enables task transferability by unifying the generation processes of different downstream tasks into the masking and recovery of modalities and sub-trajectories. Extensive experiments on two real-world vehicle trajectory datasets and three representative tasks prove that TrajFM meets its design goal.

\newpage
\appendix

\begin{table*}
    \caption{Performance of methods under \textit{wo ft} setting on trajectory arrival time estimation.}
    \label{tab:woft-trajectory-eta}
\begin{threeparttable}
\begin{tabular}{c|ccc|ccc}
\toprule
Dataset & \multicolumn{3}{c|}{Chengdu} & \multicolumn{3}{c}{Xian} \\
\midrule
\multirow{2}{*}{\diagbox[]{Method}{Metric}} & RMSE $\downarrow$ & MAE $\downarrow$ & MAPE $\downarrow$ & RMSE $\downarrow$ & MAE $\downarrow$ & MAPE $\downarrow$ \\
& (seconds) & (seconds) & (\%) & (seconds) & (seconds) & (\%) \\
\midrule
t2vec (wo ft)  & 154.48$\pm$0.42 &  99.39$\pm$1.45 & 26.08$\pm$1.08 & 234.07$\pm$0.73 & 148.81$\pm$2.81 & 20.46$\pm$0.88 \\
Trembr (wo ft)  & 179.66$\pm$0.49 &  127.40$\pm$0.20 & 36.50$\pm$0.41 & 368.34$\pm$3.14 & 277.85$\pm$2.77 & 42.93$\pm$6.57 \\
CTLE  (wo ft)  & 166.59$\pm$3.96 &  120.73$\pm$4.05 & 30.44$\pm$1.45 & 207.29$\pm$1.91 & 119.32$\pm$4.84 & 24.66$\pm$1.38 \\
Toast (wo ft)  & 136.50$\pm$1.34 &  \underline{62.04$\pm$5.92} & \underline{14.65$\pm$3.25} & 209.14$\pm$4.16 & 111.50$\pm$4.48 & 16.68$\pm$1.18 \\
TrajCL (wo ft)  & 144.59$\pm$1.57 &  92.10$\pm$3.20 & 23.11$\pm$0.94 & 208.62$\pm$2.42 & 127.14$\pm$5.86 & 20.14$\pm$0.74 \\
START (wo ft)  & \underline{132.71$\pm$1.24} &  {65.88$\pm$0.54} & {16.12$\pm$0.27} & \underline{182.34$\pm$8.12} & \underline{85.23$\pm$2.96} & \underline{12.06$\pm$0.17} \\
LightPath (wo ft)  & 134.95$\pm$1.59 &  73.44$\pm$2.13 & 18.95$\pm$1.34 & 194.83$\pm$1.67 & 97.76$\pm$3.04 & 15.73$\pm$1.25 \\
\textbf{TrajFM (ours)} & \textbf{113.55$\pm$1.88} & \textbf{50.63$\pm$1.98} & \textbf{11.02$\pm$0.97} & \textbf{170.49$\pm$2.24} & \textbf{75.37$\pm$1.87} & \textbf{10.61$\pm$1.00} \\
\bottomrule
\end{tabular}
\begin{tablenotes}\footnotesize
\item[]{
    \textbf{Bold} denotes the best result, and \underline{underline} denotes the second-best result.
    $\downarrow$ means lower is better.
}
\end{tablenotes}
\end{threeparttable}
\end{table*}

\begin{table*}
    \caption{Performance of methods under \textit{wo ft} setting on origin-destination arrival time estimation.}
    \label{tab:woft-od-eta}
\begin{threeparttable}
\begin{tabular}{c|ccc|ccc}
\toprule
Dataset & \multicolumn{3}{c|}{Chengdu} & \multicolumn{3}{c}{Xian} \\
\midrule
\multirow{2}{*}{\diagbox[]{Method}{Metric}} & RMSE $\downarrow$ & MAE $\downarrow$ & MAPE $\downarrow$ & RMSE $\downarrow$ & MAE $\downarrow$ & MAPE $\downarrow$ \\
& (seconds) & (seconds) & (\%) & (seconds) & (seconds) & (\%) \\
\midrule

t2vec (wo ft)  & 250.82$\pm$7.51
& 195.30$\pm$5.63
& 50.06$\pm$0.56
& 398.68$\pm$0.61
& 318.53$\pm$0.57
& 71.21$\pm$0.49\\

Trembr (wo ft)   & \underline{122.35$\pm$2.50}
& \underline{73.50$\pm$2.24}
& \underline{16.45$\pm$0.62}
& 387.25$\pm$2.88
& 306.35$\pm$2.66
& 61.13$\pm$0.11 \\

CTLE  (wo ft)  & 218.50$\pm$4.73
& 157.02$\pm$3.15
& 34.89$\pm$1.12
& \underline{311.16$\pm$3.18}
& \underline{223.36$\pm$3.15}
& \underline{51.39$\pm$1.82} \\

Toast (wo ft)   & 229.49$\pm$5.73
& 177.84$\pm$4.19
& 49.96$\pm$1.08
& 394.46$\pm$4.98
& 310.69$\pm$3.43
& 62.64$\pm$1.74\\

TrajCL (wo ft)   & 241.29$\pm$1.98
& 186.22$\pm$1.36
& 53.53$\pm$0.70
& 392.66$\pm$1.17
& 313.88$\pm$0.94
& 71.76$\pm$0.61\\

START (wo ft)  & 243.28$\pm$5.17
& 186.61$\pm$4.43
& 49.42$\pm$0.41
& 392.80$\pm$1.28
& 313.19$\pm$1.31
& 71.42$\pm$0.46 \\

LightPath (wo ft)  & 232.02$\pm$3.40 &  180.92$\pm$2.61 & 52.91$\pm$0.93 & 397.78$\pm$4.63 & 318.54 $\pm$ 4.43 & 70.92 $\pm$ 0.83 \\

\textbf{TrajFM (ours)} & \textbf{94.28$\pm$6.10} & \textbf{47.96$\pm$4.81} & \textbf{11.09$\pm$1.09} & \textbf{193.61$\pm$6.44} & \textbf{102.02$\pm$0.70} & \textbf{12.92$\pm$1.11} \\
\bottomrule
\end{tabular}
\begin{tablenotes}\footnotesize
\item[]{
    \textbf{Bold} denotes the best result, and \underline{underline} denotes the second-best result.
    $\downarrow$ means lower is better.
}
\end{tablenotes}
\end{threeparttable}
\end{table*}

\begin{table}
    \caption{Performance of \textit{wo ft} variants on trajectory prediction.}
    \label{tab:woft-trajectory-prediction}
    \resizebox{1.0\linewidth}{!}{
\begin{threeparttable}
\begin{tabular}{c|cc|cc}
\toprule
Dataset & \multicolumn{2}{c|}{Chengdu} & \multicolumn{2}{c}{Xian} \\
\midrule
\multirow{2}{*}{\diagbox[]{Method}{Metric}} & RMSE $\downarrow$ & MAE $\downarrow$ & RMSE $\downarrow$ & MAE $\downarrow$ \\
& (meters) & (meters) & (meters) & (meters) \\
\midrule
t2vec (wo ft) & 2329.63$\pm$21.09 & 1868.49$\pm$19.49 & 2582.14$\pm$46.79 & 2235.27$\pm$39.44\\
Trembr (wo ft) & 1787.18$\pm$29.01 & 1419.58$\pm$28.95 & 2067.80$\pm$16.30 & 1749.76$\pm$18.82\\
CTLE  (wo ft) & 3421.09$\pm$17.10 & 3041.49$\pm$23.49 & 3548.88$\pm$4.27 & 3320.46$\pm$1.12\\
Toast (wo ft) & 3434.84$\pm$9.55 & 3061.91$\pm$14.99 & 3549.65$\pm$6.42 & 3325.48$\pm$8.21\\
TrajCL (wo ft) & \underline{1059.81$\pm$16.22} & \underline{865.48$\pm$10.60} & \underline{1268.41$\pm$19.57} & \underline{1054.21$\pm$18.54}\\
START (wo ft) & 1347.13$\pm$30.72 & 1111.77$\pm$29.11 & 1406.06$\pm$18.42 & 1173.62$\pm$17.18\\
LightPath (wo ft)  & 2365.87$\pm$57.52 &  1948.97$\pm$57.78 & 2177.27$\pm$60.03 & 1859.35$\pm$48.50 \\
\textbf{TrajFM (ours)} & \textbf{196.10$\pm$20.73} & \textbf{150.86$\pm$23.31} & \textbf{320.32$\pm$9.81} &  \textbf{207.89$\pm$5.61} \\
\bottomrule
\end{tabular}
\begin{tablenotes}\footnotesize
\item[]{
    \textbf{Bold} denotes the best result, and \underline{underline} denotes the second-best result.
    $\downarrow$ means lower is better.
}
\end{tablenotes}
\end{threeparttable}
}
\end{table}

\section{Performance under \textit{wo ft} Setting}
\label{sec:wo-ft-variants}
Tables~\ref{tab:woft-trajectory-eta} to~\ref{tab:woft-trajectory-prediction} present the full results of the performance of comparison methods in the \textit{wo ft} setting and TrajFM. As observed in Section~\ref{sec:overall-performance}, comparison methods experience significant performance degradation when their parameters are not fine-tuned for the task. In other words, TrajFM shows clear performance superiority when compared to these methods in the \textit{wo ft} setting. Note that comparison methods in this setting still require training of parameters in prediction modules for task adaptation, while TrajFM does not require any additional training. This highlights the superior task transferability of TrajFM.

\section{Model Analysis}
We perform the analysis on the modules and hyper-parameters of TrajFM using the Chengdu dataset for the trajectory prediction task.

\subsection{Effectiveness of Modules}
\label{sec:modules}
We compare the \textit{full} version of TrajFM with the following variants:
\begin{enumerate}[leftmargin=*]
    \item \textit{w/o STRPE}: Replaces the STRPE with a vanilla Transformer encoder.
    \item \textit{w/o POI}: Removes the POI modality.
    \item \textit{w/o pre-train}: Skips the pre-training process and trained the model with task supervision. 
\end{enumerate}

Table~\ref{tab:ablation} compares the results. It can be observed that the vanilla Transformer encoder performs worse than the proposed STRPE, highlighting STRPE’s superiority in modeling trajectories. Removing the POI modality negatively impacts performance, indicating its contribution to TrajFM's effectiveness. Finally, training the model with task supervision instead of following the pre-training process produces worse results, demonstrating the effectiveness of the proposed pre-training process. Additionally, training the model with task supervision reduces TrajFM's task transferability.

\begin{table}
    \caption{Effectiveness of modules.}
    \label{tab:ablation}
    \begin{threeparttable}
\begin{tabular}{c|cc}
\toprule
\multirow{2}{*}{\diagbox[]{Variant}{Metric}} & RMSE $\downarrow$ & MAE $\downarrow$ \\
& (meters) & (meters) \\
\midrule
w/o STRPE & 202.83$\pm$22.97 & 156.26$\pm$18.02 \\
w/o POI & 215.53$\pm$17.39  & 169.54$\pm$15.84 \\
w/o pre-train & 221.47$\pm$13.04 & 173.72$\pm$12.49 \\
full & 196.10$\pm$20.73 & 150.86$\pm$23.31 \\
\bottomrule
\end{tabular}
\end{threeparttable}
\end{table}

\begin{figure}
    \centering
    \pgfplotstableread[row sep=\\,col sep=&]{
v & mae & rmse \\
1 & 173.42 & 210.39 \\
2 & 150.86 & 196.1 \\
3 & 164.55 & 200.37 \\
4 & 178.84 & 220.65 \\
5 & 219.21 & 274.36 \\
6 & 243.74 & 325.46 \\
}\NumLayers

\pgfplotstableread[row sep=\\,col sep=&]{
v & mae & rmse \\
32 & 160.41 & 224.05 \\
64 & 152.72 & 200.23 \\
128 & 150.86 & 196.1 \\
256 & 151.04 & 196.23 \\
512 & 154.97 & 198.19 \\
}\ModelDimD

\newcommand{\maeMin}{130}
\newcommand{\maeMax}{275}
\newcommand{\maeTick}{20}
\newcommand{\rmseMin}{190}
\newcommand{\rmseMax}{335}
\newcommand{\rmseTick}{20}

\tikzset{every plot/.style={line width=1.5pt}}

\ref{fig:shared-legend}

\begin{subfigure}[b]{0.48\linewidth}
    \begin{tikzpicture}
    \begin{axis}[
        width=1.15\linewidth, height=1.0\linewidth,
        yticklabel style = {mae},
        ymin=\maeMin, ymax=\maeMax,
        ytick distance={\maeTick},
        xtick=data,
        symbolic x coords={1,2,3,4,5,6},
        legend entries={MSE (meters), RMSE (meters)},
        legend to name=fig:shared-legend,
        legend columns=-1,
    ]
    \addplot[color=mae,mark=square*] table[x=v,y=mae]{\NumLayers};
    \addlegendentry{MSE (meters)}
    \addlegendimage{/pgfplots/refstyle=rmseStyle}
    \addlegendentry{RMSE (meters)}
    \end{axis}
    \begin{axis}[
        width=1.15\linewidth, height=1.0\linewidth,
        yticklabel style = {rmse},
        axis y line*=right,
        axis x line=none, 
        ylabel near ticks,
        ymin=\rmseMin, ymax=\rmseMax,
        ytick distance={\rmseTick},
        xtick=data, 
        symbolic x coords={1,2,3,4,5,6},
    ]
    \addplot[color=rmse,mark=triangle*] table[x=v,y=rmse]{\NumLayers};
    \label{rmseStyle}
    \end{axis}
    \end{tikzpicture}
    \caption{Number of Layers $L$}
    \label{fig:num-layers}
\end{subfigure}
\hfill
\begin{subfigure}[b]{0.48\linewidth}
    \begin{tikzpicture}
    \begin{axis}[
        width=1.15\linewidth, height=1.0\linewidth,
        yticklabel style = {mae},
        ymin=\maeMin, ymax=\maeMax,
        ytick distance={\maeTick},
        xtick=data,
        symbolic x coords={32,64,128,256,512},
    ]
    \addplot[color=mae,mark=square*] table[x=v,y=mae]{\ModelDimD};
    \end{axis}
    \begin{axis}[
        width=1.15\linewidth, height=1.0\linewidth,
        yticklabel style = {rmse},
        axis y line*=right,
        axis x line=none, 
        ylabel near ticks,
        ymin=\rmseMin, ymax=\rmseMax,
        ytick distance={\rmseTick},
        xtick=data, 
        symbolic x coords={32,64,128,256,512},
    ]
    \addplot[color=rmse,mark=triangle*] table[x=v,y=rmse]{\ModelDimD};
    \end{axis}
    \end{tikzpicture}
    \caption{Model Dimension $d$}
    \label{fig:model-dim}
\end{subfigure}
    \caption{Effectiveness of hyper-parameters.}
    \label{fig:hyper-parameter}
\end{figure}
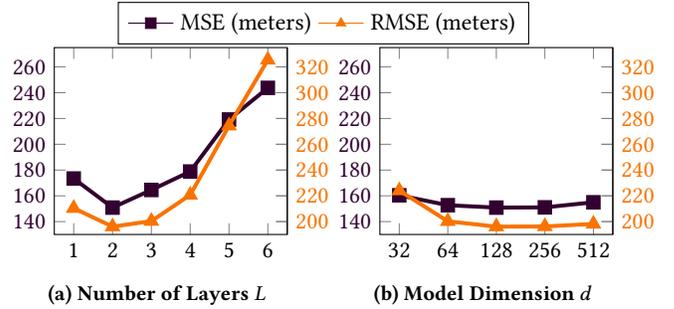

\subsection{Effectiveness of Hyper-parameters}
\label{sec:hyper-parameters}
Figure~\ref{fig:hyper-parameter} illustrates the effectiveness of the two key hyper-parameters. Both the number of layers $L$ and the model dimension $d$ control the model capacity. We observe that $L$ has an optimal value of 2; beyond that, the model becomes overly complicated and harder to train. $d$ reaches its optimal value at 128. A smaller value reduces the model's learning capacity, while a larger value offers little improvement in performance and worsens computational efficiency.

\bibliographystyle{ACM-Reference-Format}
\bibliography{reference}


\begin{thebibliography}{44}


\ifx \showCODEN    \undefined \def \showCODEN     #1{\unskip}     \fi
\ifx \showDOI      \undefined \def \showDOI       #1{#1}\fi
\ifx \showISBNx    \undefined \def \showISBNx     #1{\unskip}     \fi
\ifx \showISBNxiii \undefined \def \showISBNxiii  #1{\unskip}     \fi
\ifx \showISSN     \undefined \def \showISSN      #1{\unskip}     \fi
\ifx \showLCCN     \undefined \def \showLCCN      #1{\unskip}     \fi
\ifx \shownote     \undefined \def \shownote      #1{#1}          \fi
\ifx \showarticletitle \undefined \def \showarticletitle #1{#1}   \fi
\ifx \showURL      \undefined \def \showURL       {\relax}        \fi
\providecommand\bibfield[2]{#2}
\providecommand\bibinfo[2]{#2}
\providecommand\natexlab[1]{#1}
\providecommand\showeprint[2][]{arXiv:#2}

\bibitem[Chang et~al\mbox{.}(2023)]%
        {DBLP:conf/icde/Chang0LT23}
\bibfield{author}{\bibinfo{person}{Yanchuan Chang}, \bibinfo{person}{Jianzhong Qi}, \bibinfo{person}{Yuxuan Liang}, {and} \bibinfo{person}{Egemen Tanin}.} \bibinfo{year}{2023}\natexlab{}.
\newblock \showarticletitle{Contrastive Trajectory Similarity Learning with Dual-Feature Attention}. In \bibinfo{booktitle}{\emph{ICDE}}. \bibinfo{pages}{2933--2945}.
\newblock


\bibitem[Chen et~al\mbox{.}(2020)]%
        {DBLP:conf/icml/ChenK0H20}
\bibfield{author}{\bibinfo{person}{Ting Chen}, \bibinfo{person}{Simon Kornblith}, \bibinfo{person}{Mohammad Norouzi}, {and} \bibinfo{person}{Geoffrey~E. Hinton}.} \bibinfo{year}{2020}\natexlab{}.
\newblock \showarticletitle{A Simple Framework for Contrastive Learning of Visual Representations}. In \bibinfo{booktitle}{\emph{ICML}}, Vol.~\bibinfo{volume}{119}. \bibinfo{pages}{1597--1607}.
\newblock


\bibitem[Chen et~al\mbox{.}(2024)]%
        {DBLP:journals/tkdd/ChenHYJD24}
\bibfield{author}{\bibinfo{person}{Wei Chen}, \bibinfo{person}{Chao Huang}, \bibinfo{person}{Yanwei Yu}, \bibinfo{person}{Yongguo Jiang}, {and} \bibinfo{person}{Junyu Dong}.} \bibinfo{year}{2024}\natexlab{}.
\newblock \showarticletitle{Trajectory-User Linking via Hierarchical Spatio-Temporal Attention Networks}.
\newblock \bibinfo{journal}{\emph{{ACM} Trans. Knowl. Discov. Data}} \bibinfo{volume}{18}, \bibinfo{number}{4} (\bibinfo{year}{2024}), \bibinfo{pages}{85:1--85:22}.
\newblock


\bibitem[Chen et~al\mbox{.}(2021)]%
        {DBLP:conf/cikm/ChenLCBLLCE21}
\bibfield{author}{\bibinfo{person}{Yile Chen}, \bibinfo{person}{Xiucheng Li}, \bibinfo{person}{Gao Cong}, \bibinfo{person}{Zhifeng Bao}, \bibinfo{person}{Cheng Long}, \bibinfo{person}{Yiding Liu}, \bibinfo{person}{Arun~Kumar Chandran}, {and} \bibinfo{person}{Richard Ellison}.} \bibinfo{year}{2021}\natexlab{}.
\newblock \showarticletitle{Robust Road Network Representation Learning: When Traffic Patterns Meet Traveling Semantics}. In \bibinfo{booktitle}{\emph{CIKM}}. \bibinfo{pages}{211--220}.
\newblock


\bibitem[Chung et~al\mbox{.}(2014)]%
        {DBLP:journals/corr/ChungGCB14}
\bibfield{author}{\bibinfo{person}{Junyoung Chung}, \bibinfo{person}{Caglar Gulcehre}, \bibinfo{person}{KyungHyun Cho}, {and} \bibinfo{person}{Yoshua Bengio}.} \bibinfo{year}{2014}\natexlab{}.
\newblock \showarticletitle{Empirical evaluation of gated recurrent neural networks on sequence modeling}.
\newblock \bibinfo{journal}{\emph{arXiv preprint arXiv:1412.3555}} (\bibinfo{year}{2014}).
\newblock


\bibitem[Devlin et~al\mbox{.}(2019)]%
        {DBLP:conf/naacl/DevlinCLT19}
\bibfield{author}{\bibinfo{person}{Jacob Devlin}, \bibinfo{person}{Ming{-}Wei Chang}, \bibinfo{person}{Kenton Lee}, {and} \bibinfo{person}{Kristina Toutanova}.} \bibinfo{year}{2019}\natexlab{}.
\newblock \showarticletitle{{BERT:} Pre-training of Deep Bidirectional Transformers for Language Understanding}. In \bibinfo{booktitle}{\emph{NAACL}}. \bibinfo{pages}{4171--4186}.
\newblock


\bibitem[Du et~al\mbox{.}(2022)]%
        {DBLP:conf/acl/DuQLDQY022}
\bibfield{author}{\bibinfo{person}{Zhengxiao Du}, \bibinfo{person}{Yujie Qian}, \bibinfo{person}{Xiao Liu}, \bibinfo{person}{Ming Ding}, \bibinfo{person}{Jiezhong Qiu}, \bibinfo{person}{Zhilin Yang}, {and} \bibinfo{person}{Jie Tang}.} \bibinfo{year}{2022}\natexlab{}.
\newblock \showarticletitle{{GLM:} {G}eneral Language Model Pretraining with Autoregressive Blank Infilling}. In \bibinfo{booktitle}{\emph{ACL}}. \bibinfo{pages}{320--335}.
\newblock


\bibitem[Fang et~al\mbox{.}(2022)]%
        {DBLP:conf/kdd/Fang0ZHCGJ22}
\bibfield{author}{\bibinfo{person}{Ziquan Fang}, \bibinfo{person}{Yuntao Du}, \bibinfo{person}{Xinjun Zhu}, \bibinfo{person}{Danlei Hu}, \bibinfo{person}{Lu Chen}, \bibinfo{person}{Yunjun Gao}, {and} \bibinfo{person}{Christian~S. Jensen}.} \bibinfo{year}{2022}\natexlab{}.
\newblock \showarticletitle{Spatio-Temporal Trajectory Similarity Learning in Road Networks}. In \bibinfo{booktitle}{\emph{KDD}}. \bibinfo{pages}{347--356}.
\newblock


\bibitem[Feng et~al\mbox{.}(2018)]%
        {DBLP:conf/www/FengLZSMGJ18}
\bibfield{author}{\bibinfo{person}{Jie Feng}, \bibinfo{person}{Yong Li}, \bibinfo{person}{Chao Zhang}, \bibinfo{person}{Funing Sun}, \bibinfo{person}{Fanchao Meng}, \bibinfo{person}{Ang Guo}, {and} \bibinfo{person}{Depeng Jin}.} \bibinfo{year}{2018}\natexlab{}.
\newblock \showarticletitle{DeepMove: Predicting Human Mobility with Attentional Recurrent Networks}. In \bibinfo{booktitle}{\emph{WWW}}. \bibinfo{pages}{1459--1468}.
\newblock


\bibitem[Fu and Lee(2020)]%
        {DBLP:journals/tist/FuL20}
\bibfield{author}{\bibinfo{person}{Tao{-}Yang Fu} {and} \bibinfo{person}{Wang{-}Chien Lee}.} \bibinfo{year}{2020}\natexlab{}.
\newblock \showarticletitle{{TremBR}: {E}xploring Road Networks for Trajectory Representation Learning}.
\newblock \bibinfo{journal}{\emph{{ACM} Trans. Intell. Syst. Technol.}} \bibinfo{volume}{11}, \bibinfo{number}{1} (\bibinfo{year}{2020}), \bibinfo{pages}{10:1--10:25}.
\newblock


\bibitem[Gan et~al\mbox{.}(2021)]%
        {DBLP:conf/gis/GanZW21}
\bibfield{author}{\bibinfo{person}{Yunchong Gan}, \bibinfo{person}{Haoyu Zhang}, {and} \bibinfo{person}{Mingjie Wang}.} \bibinfo{year}{2021}\natexlab{}.
\newblock \showarticletitle{Travel Time Estimation Based on Neural Network with Auxiliary Loss}. In \bibinfo{booktitle}{\emph{SIGSPATIAL}}. \bibinfo{pages}{642--645}.
\newblock


\bibitem[Han et~al\mbox{.}(2022)]%
        {DBLP:journals/pvldb/HanCMG22}
\bibfield{author}{\bibinfo{person}{Xiaolin Han}, \bibinfo{person}{Reynold Cheng}, \bibinfo{person}{Chenhao Ma}, {and} \bibinfo{person}{Tobias Grubenmann}.} \bibinfo{year}{2022}\natexlab{}.
\newblock \showarticletitle{{DeepTEA}: {E}ffective and Efficient Online Time-dependent Trajectory Outlier Detection}.
\newblock \bibinfo{journal}{\emph{PVLDB}} \bibinfo{volume}{15}, \bibinfo{number}{7} (\bibinfo{year}{2022}), \bibinfo{pages}{1493--1505}.
\newblock


\bibitem[Hinton and Salakhutdinov(2006)]%
        {hinton2006reducing}
\bibfield{author}{\bibinfo{person}{Geoffrey~E Hinton} {and} \bibinfo{person}{Ruslan~R Salakhutdinov}.} \bibinfo{year}{2006}\natexlab{}.
\newblock \showarticletitle{Reducing the dimensionality of data with neural networks}.
\newblock \bibinfo{journal}{\emph{science}} \bibinfo{volume}{313}, \bibinfo{number}{5786} (\bibinfo{year}{2006}), \bibinfo{pages}{504--507}.
\newblock


\bibitem[Hochreiter and Schmidhuber(1997)]%
        {hochreiter1997long}
\bibfield{author}{\bibinfo{person}{Sepp Hochreiter} {and} \bibinfo{person}{J{\"u}rgen Schmidhuber}.} \bibinfo{year}{1997}\natexlab{}.
\newblock \showarticletitle{Long short-term memory}.
\newblock \bibinfo{journal}{\emph{Neural computation}} \bibinfo{volume}{9}, \bibinfo{number}{8} (\bibinfo{year}{1997}), \bibinfo{pages}{1735--1780}.
\newblock


\bibitem[Jiang et~al\mbox{.}(2023)]%
        {DBLP:conf/icde/JiangPRJLW23}
\bibfield{author}{\bibinfo{person}{Jiawei Jiang}, \bibinfo{person}{Dayan Pan}, \bibinfo{person}{Houxing Ren}, \bibinfo{person}{Xiaohan Jiang}, \bibinfo{person}{Chao Li}, {and} \bibinfo{person}{Jingyuan Wang}.} \bibinfo{year}{2023}\natexlab{}.
\newblock \showarticletitle{Self-supervised Trajectory Representation Learning with Temporal Regularities and Travel Semantics}. In \bibinfo{booktitle}{\emph{ICDE}}. \bibinfo{pages}{843--855}.
\newblock


\bibitem[Kong and Wu(2018)]%
        {DBLP:conf/ijcai/Kong018}
\bibfield{author}{\bibinfo{person}{Dejiang Kong} {and} \bibinfo{person}{Fei Wu}.} \bibinfo{year}{2018}\natexlab{}.
\newblock \showarticletitle{{HST-LSTM:} {A} Hierarchical Spatial-Temporal Long-Short Term Memory Network for Location Prediction}. In \bibinfo{booktitle}{\emph{IJCAI}}. \bibinfo{pages}{2341--2347}.
\newblock


\bibitem[Li et~al\mbox{.}({[n.\,d.]})]%
        {DBLP:conf/icde/LiZCJW18}
\bibfield{author}{\bibinfo{person}{Xiucheng Li}, \bibinfo{person}{Kaiqi Zhao}, \bibinfo{person}{Gao Cong}, \bibinfo{person}{Christian~S. Jensen}, {and} \bibinfo{person}{Wei Wei}.} \bibinfo{year}{[n.\,d.]}\natexlab{}.
\newblock \showarticletitle{Deep Representation Learning for Trajectory Similarity Computation}. In \bibinfo{booktitle}{\emph{ICDE}}. \bibinfo{pages}{617--628}.
\newblock


\bibitem[Li et~al\mbox{.}(2018)]%
        {DBLP:conf/kdd/LiFWSYL18}
\bibfield{author}{\bibinfo{person}{Yaguang Li}, \bibinfo{person}{Kun Fu}, \bibinfo{person}{Zheng Wang}, \bibinfo{person}{Cyrus Shahabi}, \bibinfo{person}{Jieping Ye}, {and} \bibinfo{person}{Yan Liu}.} \bibinfo{year}{2018}\natexlab{}.
\newblock \showarticletitle{Multi-task Representation Learning for Travel Time Estimation}. In \bibinfo{booktitle}{\emph{KDD}}. \bibinfo{pages}{1695--1704}.
\newblock


\bibitem[Liang et~al\mbox{.}(2022)]%
        {DBLP:conf/cikm/LiangOWLCZZZ22}
\bibfield{author}{\bibinfo{person}{Yuxuan Liang}, \bibinfo{person}{Kun Ouyang}, \bibinfo{person}{Yiwei Wang}, \bibinfo{person}{Xu Liu}, \bibinfo{person}{Hongyang Chen}, \bibinfo{person}{Junbo Zhang}, \bibinfo{person}{Yu Zheng}, {and} \bibinfo{person}{Roger Zimmermann}.} \bibinfo{year}{2022}\natexlab{}.
\newblock \showarticletitle{TrajFormer: Efficient Trajectory Classification with Transformers}. In \bibinfo{booktitle}{\emph{CIKM}}. \bibinfo{pages}{1229--1237}.
\newblock


\bibitem[Liang et~al\mbox{.}(2021)]%
        {DBLP:conf/ijcai/LiangOYWTZ21}
\bibfield{author}{\bibinfo{person}{Yuxuan Liang}, \bibinfo{person}{Kun Ouyang}, \bibinfo{person}{Hanshu Yan}, \bibinfo{person}{Yiwei Wang}, \bibinfo{person}{Zekun Tong}, {and} \bibinfo{person}{Roger Zimmermann}.} \bibinfo{year}{2021}\natexlab{}.
\newblock \showarticletitle{Modeling Trajectories with Neural Ordinary Differential Equations}. In \bibinfo{booktitle}{\emph{IJCAI}}. \bibinfo{pages}{1498--1504}.
\newblock


\bibitem[Lin et~al\mbox{.}(2023a)]%
        {10375102}
\bibfield{author}{\bibinfo{person}{Yan Lin}, \bibinfo{person}{Huaiyu Wan}, \bibinfo{person}{Shengnan Guo}, \bibinfo{person}{Jilin Hu}, \bibinfo{person}{Christian~S. Jensen}, {and} \bibinfo{person}{Youfang Lin}.} \bibinfo{year}{2023}\natexlab{a}.
\newblock \showarticletitle{Pre-Training General Trajectory Embeddings With Maximum Multi-View Entropy Coding}.
\newblock \bibinfo{journal}{\emph{{IEEE} Trans. Knowl. Data Eng.}} (\bibinfo{year}{2023}), \bibinfo{pages}{1--15}.
\newblock


\bibitem[Lin et~al\mbox{.}(2021)]%
        {DBLP:conf/aaai/LinW0L21}
\bibfield{author}{\bibinfo{person}{Yan Lin}, \bibinfo{person}{Huaiyu Wan}, \bibinfo{person}{Shengnan Guo}, {and} \bibinfo{person}{Youfang Lin}.} \bibinfo{year}{2021}\natexlab{}.
\newblock \showarticletitle{Pre-training Context and Time Aware Location Embeddings from Spatial-Temporal Trajectories for User Next Location Prediction}. In \bibinfo{booktitle}{\emph{AAAI}}. \bibinfo{pages}{4241--4248}.
\newblock


\bibitem[Lin et~al\mbox{.}(2023b)]%
        {DBLP:journals/pacmmod/LinWHGYLJ23}
\bibfield{author}{\bibinfo{person}{Yan Lin}, \bibinfo{person}{Huaiyu Wan}, \bibinfo{person}{Jilin Hu}, \bibinfo{person}{Shengnan Guo}, \bibinfo{person}{Bin Yang}, \bibinfo{person}{Youfang Lin}, {and} \bibinfo{person}{Christian~S. Jensen}.} \bibinfo{year}{2023}\natexlab{b}.
\newblock \showarticletitle{Origin-Destination Travel Time Oracle for Map-based Services}.
\newblock \bibinfo{journal}{\emph{PACMMOD}} \bibinfo{volume}{1}, \bibinfo{number}{3} (\bibinfo{year}{2023}), \bibinfo{pages}{217:1--217:27}.
\newblock


\bibitem[Liu et~al\mbox{.}(2020)]%
        {DBLP:conf/icde/Liu0CB20}
\bibfield{author}{\bibinfo{person}{Yiding Liu}, \bibinfo{person}{Kaiqi Zhao}, \bibinfo{person}{Gao Cong}, {and} \bibinfo{person}{Zhifeng Bao}.} \bibinfo{year}{2020}\natexlab{}.
\newblock \showarticletitle{Online Anomalous Trajectory Detection with Deep Generative Sequence Modeling}. In \bibinfo{booktitle}{\emph{ICDE}}. \bibinfo{pages}{949--960}.
\newblock


\bibitem[Miao et~al\mbox{.}(2020)]%
        {DBLP:conf/wsdm/MiaoLZW20}
\bibfield{author}{\bibinfo{person}{Congcong Miao}, \bibinfo{person}{Ziyan Luo}, \bibinfo{person}{Fengzhu Zeng}, {and} \bibinfo{person}{Jilong Wang}.} \bibinfo{year}{2020}\natexlab{}.
\newblock \showarticletitle{Predicting Human Mobility via Attentive Convolutional Network}. In \bibinfo{booktitle}{\emph{WSDM}}. \bibinfo{pages}{438--446}.
\newblock


\bibitem[Paszke et~al\mbox{.}(2019)]%
        {DBLP:conf/nips/PaszkeGMLBCKLGA19}
\bibfield{author}{\bibinfo{person}{Adam Paszke}, \bibinfo{person}{Sam Gross}, \bibinfo{person}{Francisco Massa}, \bibinfo{person}{Adam Lerer}, \bibinfo{person}{James Bradbury}, \bibinfo{person}{Gregory Chanan}, \bibinfo{person}{Trevor Killeen}, \bibinfo{person}{Zeming Lin}, \bibinfo{person}{Natalia Gimelshein}, \bibinfo{person}{Luca Antiga}, \bibinfo{person}{Alban Desmaison}, \bibinfo{person}{Andreas K{\"{o}}pf}, \bibinfo{person}{Edward~Z. Yang}, \bibinfo{person}{Zachary DeVito}, \bibinfo{person}{Martin Raison}, \bibinfo{person}{Alykhan Tejani}, \bibinfo{person}{Sasank Chilamkurthy}, \bibinfo{person}{Benoit Steiner}, \bibinfo{person}{Lu Fang}, \bibinfo{person}{Junjie Bai}, {and} \bibinfo{person}{Soumith Chintala}.} \bibinfo{year}{2019}\natexlab{}.
\newblock \showarticletitle{{PyTorch}: {A}n Imperative Style, High-Performance Deep Learning Library}. In \bibinfo{booktitle}{\emph{NeurIPS}}. \bibinfo{pages}{8024--8035}.
\newblock


\bibitem[Radford et~al\mbox{.}(2021)]%
        {DBLP:conf/icml/RadfordKHRGASAM21}
\bibfield{author}{\bibinfo{person}{Alec Radford}, \bibinfo{person}{Jong~Wook Kim}, \bibinfo{person}{Chris Hallacy}, \bibinfo{person}{Aditya Ramesh}, \bibinfo{person}{Gabriel Goh}, \bibinfo{person}{Sandhini Agarwal}, \bibinfo{person}{Girish Sastry}, \bibinfo{person}{Amanda Askell}, \bibinfo{person}{Pamela Mishkin}, \bibinfo{person}{Jack Clark}, \bibinfo{person}{Gretchen Krueger}, {and} \bibinfo{person}{Ilya Sutskever}.} \bibinfo{year}{2021}\natexlab{}.
\newblock \showarticletitle{Learning Transferable Visual Models From Natural Language Supervision}. In \bibinfo{booktitle}{\emph{ICML}}, Vol.~\bibinfo{volume}{139}. \bibinfo{pages}{8748--8763}.
\newblock


\bibitem[Sang et~al\mbox{.}(2023)]%
        {DBLP:journals/www/SangXCZ23}
\bibfield{author}{\bibinfo{person}{Yu Sang}, \bibinfo{person}{Zhenping Xie}, \bibinfo{person}{Wei Chen}, {and} \bibinfo{person}{Lei Zhao}.} \bibinfo{year}{2023}\natexlab{}.
\newblock \showarticletitle{{TULRN:} Trajectory user linking on road networks}.
\newblock \bibinfo{journal}{\emph{WWW}} \bibinfo{volume}{26}, \bibinfo{number}{4} (\bibinfo{year}{2023}), \bibinfo{pages}{1949--1965}.
\newblock


\bibitem[Su et~al\mbox{.}(2024)]%
        {DBLP:journals/ijon/SuALPBL24}
\bibfield{author}{\bibinfo{person}{Jianlin Su}, \bibinfo{person}{Murtadha H.~M. Ahmed}, \bibinfo{person}{Yu Lu}, \bibinfo{person}{Shengfeng Pan}, \bibinfo{person}{Wen Bo}, {and} \bibinfo{person}{Yunfeng Liu}.} \bibinfo{year}{2024}\natexlab{}.
\newblock \showarticletitle{{RoFormer}: {E}nhanced transformer with Rotary Position Embedding}.
\newblock \bibinfo{journal}{\emph{Neurocomputing}}  \bibinfo{volume}{568} (\bibinfo{year}{2024}), \bibinfo{pages}{127063}.
\newblock


\bibitem[Tancik et~al\mbox{.}(2020)]%
        {DBLP:conf/nips/TancikSMFRSRBN20}
\bibfield{author}{\bibinfo{person}{Matthew Tancik}, \bibinfo{person}{Pratul~P. Srinivasan}, \bibinfo{person}{Ben Mildenhall}, \bibinfo{person}{Sara Fridovich{-}Keil}, \bibinfo{person}{Nithin Raghavan}, \bibinfo{person}{Utkarsh Singhal}, \bibinfo{person}{Ravi Ramamoorthi}, \bibinfo{person}{Jonathan~T. Barron}, {and} \bibinfo{person}{Ren Ng}.} \bibinfo{year}{2020}\natexlab{}.
\newblock \showarticletitle{Fourier Features Let Networks Learn High Frequency Functions in Low Dimensional Domains}. In \bibinfo{booktitle}{\emph{NeurIPS}}.
\newblock


\bibitem[Vaswani et~al\mbox{.}(2017)]%
        {DBLP:conf/nips/VaswaniSPUJGKP17}
\bibfield{author}{\bibinfo{person}{Ashish Vaswani}, \bibinfo{person}{Noam Shazeer}, \bibinfo{person}{Niki Parmar}, \bibinfo{person}{Jakob Uszkoreit}, \bibinfo{person}{Llion Jones}, \bibinfo{person}{Aidan~N. Gomez}, \bibinfo{person}{Lukasz Kaiser}, {and} \bibinfo{person}{Illia Polosukhin}.} \bibinfo{year}{2017}\natexlab{}.
\newblock \showarticletitle{Attention is All you Need}. In \bibinfo{booktitle}{\emph{NeurIPS}}. \bibinfo{pages}{5998--6008}.
\newblock


\bibitem[Wan et~al\mbox{.}(2022)]%
        {DBLP:journals/tkde/WanLGL22}
\bibfield{author}{\bibinfo{person}{Huaiyu Wan}, \bibinfo{person}{Yan Lin}, \bibinfo{person}{Shengnan Guo}, {and} \bibinfo{person}{Youfang Lin}.} \bibinfo{year}{2022}\natexlab{}.
\newblock \showarticletitle{Pre-Training Time-Aware Location Embeddings from Spatial-Temporal Trajectories}.
\newblock \bibinfo{journal}{\emph{{IEEE} Trans. Knowl. Data Eng.}} \bibinfo{volume}{34}, \bibinfo{number}{11} (\bibinfo{year}{2022}), \bibinfo{pages}{5510--5523}.
\newblock


\bibitem[Wang et~al\mbox{.}(2018b)]%
        {DBLP:conf/aaai/WangZCLZ18}
\bibfield{author}{\bibinfo{person}{Dong Wang}, \bibinfo{person}{Junbo Zhang}, \bibinfo{person}{Wei Cao}, \bibinfo{person}{Jian Li}, {and} \bibinfo{person}{Yu Zheng}.} \bibinfo{year}{2018}\natexlab{b}.
\newblock \showarticletitle{When Will You Arrive? {E}stimating Travel Time Based on Deep Neural Networks}. In \bibinfo{booktitle}{\emph{AAAI}}. \bibinfo{pages}{2500--2507}.
\newblock


\bibitem[Wang et~al\mbox{.}(2016)]%
        {DBLP:conf/gis/WangKKL16}
\bibfield{author}{\bibinfo{person}{Hongjian Wang}, \bibinfo{person}{Yu{-}Hsuan Kuo}, \bibinfo{person}{Daniel Kifer}, {and} \bibinfo{person}{Zhenhui Li}.} \bibinfo{year}{2016}\natexlab{}.
\newblock \showarticletitle{A simple baseline for travel time estimation using large-scale trip data}. In \bibinfo{booktitle}{\emph{SIGSPATIAL}}. \bibinfo{pages}{61:1--61:4}.
\newblock


\bibitem[Wang et~al\mbox{.}(2018a)]%
        {DBLP:conf/kdd/WangFY18}
\bibfield{author}{\bibinfo{person}{Zheng Wang}, \bibinfo{person}{Kun Fu}, {and} \bibinfo{person}{Jieping Ye}.} \bibinfo{year}{2018}\natexlab{a}.
\newblock \showarticletitle{Learning to Estimate the Travel Time}. In \bibinfo{booktitle}{\emph{KDD}}. \bibinfo{pages}{858--866}.
\newblock


\bibitem[Wei et~al\mbox{.}(2024)]%
        {10517676}
\bibfield{author}{\bibinfo{person}{Tonglong Wei}, \bibinfo{person}{Youfang Lin}, \bibinfo{person}{Yan Lin}, \bibinfo{person}{Shengnan Guo}, \bibinfo{person}{Lan Zhang}, {and} \bibinfo{person}{Huaiyu Wan}.} \bibinfo{year}{2024}\natexlab{}.
\newblock \showarticletitle{Micro-Macro Spatial-Temporal Graph-Based Encoder-Decoder for Map-Constrained Trajectory Recovery}.
\newblock \bibinfo{journal}{\emph{{IEEE} Trans. Knowl. Data Eng.}} (\bibinfo{year}{2024}), \bibinfo{pages}{1--15}.
\newblock


\bibitem[Wu and Wu(2019)]%
        {DBLP:conf/aaai/WuW19}
\bibfield{author}{\bibinfo{person}{Fan Wu} {and} \bibinfo{person}{Lixia Wu}.} \bibinfo{year}{2019}\natexlab{}.
\newblock \showarticletitle{DeepETA: {A} Spatial-Temporal Sequential Neural Network Model for Estimating Time of Arrival in Package Delivery System}. In \bibinfo{booktitle}{\emph{AAAI}}. \bibinfo{pages}{774--781}.
\newblock


\bibitem[Wu et~al\mbox{.}(2017)]%
        {DBLP:conf/ijcai/WuCSZW17}
\bibfield{author}{\bibinfo{person}{Hao Wu}, \bibinfo{person}{Ziyang Chen}, \bibinfo{person}{Weiwei Sun}, \bibinfo{person}{Baihua Zheng}, {and} \bibinfo{person}{Wei Wang}.} \bibinfo{year}{2017}\natexlab{}.
\newblock \showarticletitle{Modeling Trajectories with Recurrent Neural Networks}. In \bibinfo{booktitle}{\emph{IJCAI}}. \bibinfo{pages}{3083--3090}.
\newblock


\bibitem[Yan et~al\mbox{.}(2023)]%
        {DBLP:journals/www/YanZSYD23}
\bibfield{author}{\bibinfo{person}{Bingqi Yan}, \bibinfo{person}{Geng Zhao}, \bibinfo{person}{Lexue Song}, \bibinfo{person}{Yanwei Yu}, {and} \bibinfo{person}{Junyu Dong}.} \bibinfo{year}{2023}\natexlab{}.
\newblock \showarticletitle{{PreCLN}: {P}retrained-based contrastive learning network for vehicle trajectory prediction}.
\newblock \bibinfo{journal}{\emph{WWW}} \bibinfo{volume}{26}, \bibinfo{number}{4} (\bibinfo{year}{2023}), \bibinfo{pages}{1853--1875}.
\newblock


\bibitem[Yang et~al\mbox{.}(2023)]%
        {DBLP:conf/kdd/YangHGYJ23}
\bibfield{author}{\bibinfo{person}{Sean~Bin Yang}, \bibinfo{person}{Jilin Hu}, \bibinfo{person}{Chenjuan Guo}, \bibinfo{person}{Bin Yang}, {and} \bibinfo{person}{Christian~S. Jensen}.} \bibinfo{year}{2023}\natexlab{}.
\newblock \showarticletitle{LightPath: Lightweight and Scalable Path Representation Learning}. In \bibinfo{booktitle}{\emph{KDD}}. \bibinfo{pages}{2999--3010}.
\newblock


\bibitem[Yao et~al\mbox{.}(2017)]%
        {DBLP:conf/ijcnn/YaoZZHB17}
\bibfield{author}{\bibinfo{person}{Di Yao}, \bibinfo{person}{Chao Zhang}, \bibinfo{person}{Zhihua Zhu}, \bibinfo{person}{Jian{-}Hui Huang}, {and} \bibinfo{person}{Jingping Bi}.} \bibinfo{year}{2017}\natexlab{}.
\newblock \showarticletitle{Trajectory clustering via deep representation learning}. In \bibinfo{booktitle}{\emph{IJCNN}}. \bibinfo{pages}{3880--3887}.
\newblock


\bibitem[Yuan and Li(2021)]%
        {DBLP:journals/dase/YuanL21}
\bibfield{author}{\bibinfo{person}{Haitao Yuan} {and} \bibinfo{person}{Guoliang Li}.} \bibinfo{year}{2021}\natexlab{}.
\newblock \showarticletitle{A Survey of Traffic Prediction: from Spatio-Temporal Data to Intelligent Transportation}.
\newblock \bibinfo{journal}{\emph{Data Sci. Eng.}} \bibinfo{volume}{6}, \bibinfo{number}{1} (\bibinfo{year}{2021}), \bibinfo{pages}{63--85}.
\newblock


\bibitem[Yuan et~al\mbox{.}(2020)]%
        {DBLP:conf/sigmod/Yuan0BF20}
\bibfield{author}{\bibinfo{person}{Haitao Yuan}, \bibinfo{person}{Guoliang Li}, \bibinfo{person}{Zhifeng Bao}, {and} \bibinfo{person}{Ling Feng}.} \bibinfo{year}{2020}\natexlab{}.
\newblock \showarticletitle{Effective Travel Time Estimation: When Historical Trajectories over Road Networks Matter}. In \bibinfo{booktitle}{\emph{SIGMOD}}. \bibinfo{pages}{2135--2149}.
\newblock


\bibitem[Zhang et~al\mbox{.}(2023)]%
        {DBLP:conf/icde/Zhang0LHYLCS23}
\bibfield{author}{\bibinfo{person}{Qianru Zhang}, \bibinfo{person}{Zheng Wang}, \bibinfo{person}{Cheng Long}, \bibinfo{person}{Chao Huang}, \bibinfo{person}{Siu{-}Ming Yiu}, \bibinfo{person}{Yiding Liu}, \bibinfo{person}{Gao Cong}, {and} \bibinfo{person}{Jieming Shi}.} \bibinfo{year}{2023}\natexlab{}.
\newblock \showarticletitle{Online Anomalous Subtrajectory Detection on Road Networks with Deep Reinforcement Learning}. In \bibinfo{booktitle}{\emph{ICDE}}. \bibinfo{pages}{246--258}.
\newblock


\end{thebibliography}

\end{document}